\documentclass[journal]{IEEEtran}
\usepackage{amsmath}
\usepackage{mathrsfs}
\usepackage{subcaption}
\usepackage{epsfig}
\usepackage{graphicx}
\usepackage{algorithm}
\usepackage{algorithmic}
\usepackage{multirow}
\usepackage{ragged2e}
\usepackage{bm}

\newcommand{\etal}{\textit{et al}.}

\newcommand{\eg}{\textit{e}.\textit{g}.}

\begin{document}
%

\title{Enhancing Underexposed Photos using \\ Perceptually Bidirectional Similarity}

%
%
%


%

\author{Qing~Zhang, Yongwei Nie, Lei Zhu, Chunxia Xiao, and Wei-Shi Zheng
	\thanks{Q.~Zhang and W.-S.
		Zheng are with the School of Data and Computer Science, Sun Yat-Sen University, Guangzhou 510006, China.
		W.-S. Zheng is also with the Peng Cheng Laboratory, Shenzhen 518005, China, and the Key Laboratory of Machine Intelligence and Advanced Computing (Sun Yat-sen University), Ministry of Education, China.
		E-mail: zhangqing.whu.cs@gmail.com, wszheng@ieee.org.}
	\thanks{Y.~Nie is with the School of Computer Science and Engineering, South China University of Technology, Guangzhou 510006, China. E-mail: nieyongwei@scut.edu.cn.}
	\thanks{L.~Zhu is with the department of Computer Science and Engineering, the Chinese University of Hong Kong. E-mail: lzhu@cse.cuhk.edu.hk.}
	\thanks{ C.~Xiao is with the School of Computer Science, Wuhan University, Wuhan 430072, China. E-mail: cxxiao@whu.edu.cn.}
}

\maketitle

\begin{abstract}
Although remarkable progress has been made, existing methods for enhancing underexposed photos tend to produce visually unpleasing results due to the existence of visual artifacts (\eg, color distortion, loss of details and uneven exposure). We observed that this is because they fail to ensure the perceptual consistency of visual information between the source underexposed image and its enhanced output.
To obtain high-quality results free of these artifacts, we present a novel underexposed photo enhancement approach that is able to maintain the perceptual consistency. We achieve this by proposing an effective criterion, referred to as perceptually bidirectional similarity, which explicitly describes how to ensure the perceptual consistency. Particularly, we adopt the Retinex theory and cast the enhancement problem as a constrained illumination estimation optimization, where we formulate perceptually bidirectional similarity as constraints on illumination and solve for the illumination which can recover the desired artifact-free enhancement results.
In addition, we describe a video enhancement framework that adopts the presented illumination estimation for handling underexposed videos. To this end, a probabilistic approach is introduced to propagate illuminations of sampled keyframes to the entire video by tackling a Bayesian Maximum A Posteriori problem.
Extensive experiments demonstrate the superiority of our method over the state-of-the-art methods.
\end{abstract}


%
\IEEEpeerreviewmaketitle

\section{Introduction}
\begin{figure}
	\centering
	\begin{subfigure}[c]{0.23\textwidth}
		\centering
		\includegraphics[width=1.62in]{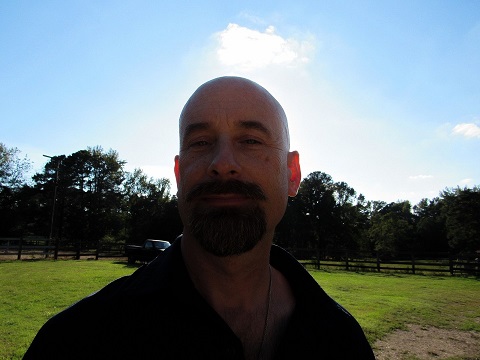}
		\caption{Input}
	\end{subfigure}
	\begin{subfigure}[c]{0.23\textwidth}
		\centering
		\includegraphics[width=1.62in]{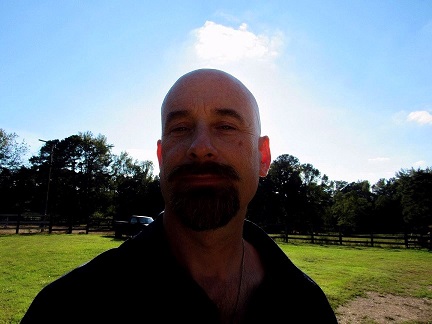}
		\caption{Auto Enhance on iPhone}
	\end{subfigure} \vspace{0.5em} \\
	\begin{subfigure}[c]{0.23\textwidth}
		\centering
		\includegraphics[width=1.62in]{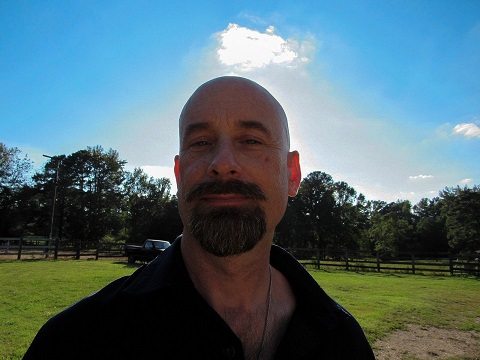}
		\caption{Auto Tone in Lightroom}
	\end{subfigure}
	\begin{subfigure}[c]{0.23\textwidth}
		\centering
		\includegraphics[width=1.62in]{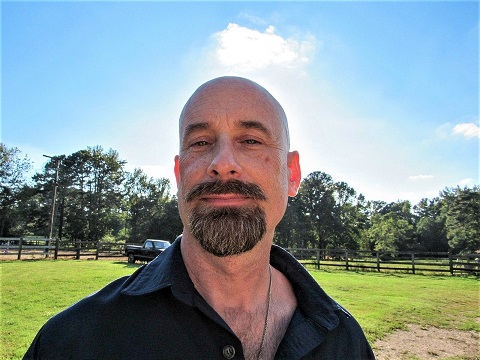}
		\caption{Our result}
	\end{subfigure}
	\caption{An example underexposed photo enhanced by existing tools and our approach. }
	\vspace{-4mm}
	\label{fig:teaser}
\end{figure}

\IEEEPARstart{W}{ith} the popularization of the readily-available cameras on cell phones, people are increasingly interested in taking photos. However, capturing well-exposed photos under complex lighting conditions (\eg, low-light and back-light) remains a challenge for non-expert users. Hence, underexposed photos are inevitably created (see Fig.~\ref{fig:teaser}(a) for an example). Due to the low detail visibility and dull colors, these photos not only look unpleasing and fail to capture what user desires, but also adversely affect various image analysis tasks, such
as segmentation \cite{liang2016clothes,kang2018depth}, object recognition \cite{nascimento2006performance,dong2016occlusion} and saliency detection \cite{li2013co,lin2018saliency}, etc. To enhance the image aesthetic and benefit subsequent applications, automatic underexposed photo enhancement techniques are thus widely required.

Underexposed photo enhancement is a challenging task, since it is highly non-linear and subjective. Commercial softwares such as Adobe Lightroom and Photoshop allow users to interactively retouch photos, while they remain difficult for non-experts. Other ease of use alternatives such as the ``Auto Enhance'' on iPhone and the ``Auto Tone'' in Lightroom allow enhancing underexposed photos by just a single click. However, they may fail to produce high-quality results due to the inherent difficulty of automatically balancing all assorted appearance factors (\eg, brightness, contrast, and saturation, etc.) in the adjustment, as shown in Fig. 1(b) and (c).

\begin{figure*}
	\centering
	\begin{subfigure}[c]{0.23\textwidth}
		\centering
		\includegraphics[width=1.65in]{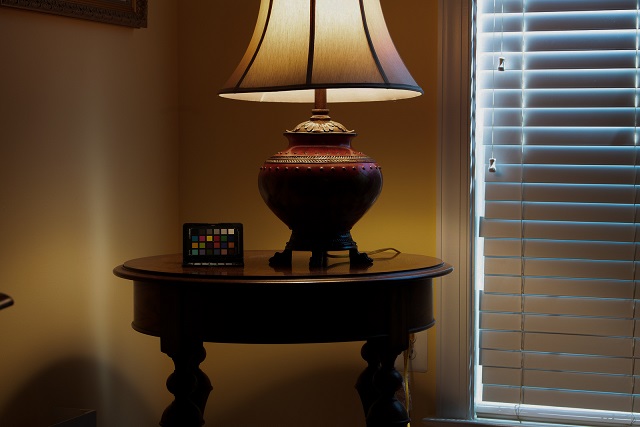}
		\caption{Input}
	\end{subfigure}
	\begin{subfigure}[c]{0.23\textwidth}
		\centering
		\includegraphics[width=1.65in]{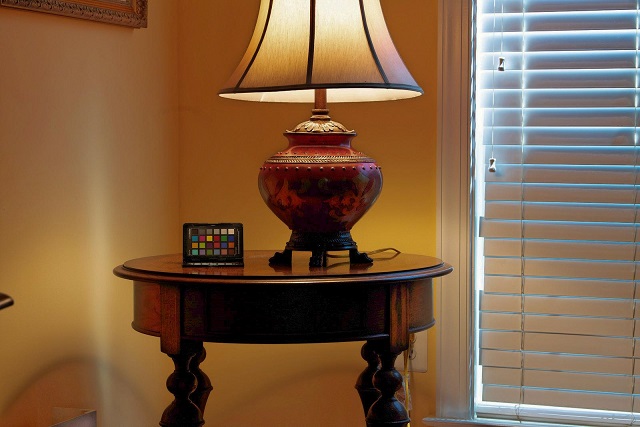}
		\caption{NPE~\cite{wang2013naturalness}}
	\end{subfigure}
	\begin{subfigure}[c]{0.23\textwidth}
		\centering
		\includegraphics[width=1.65in]{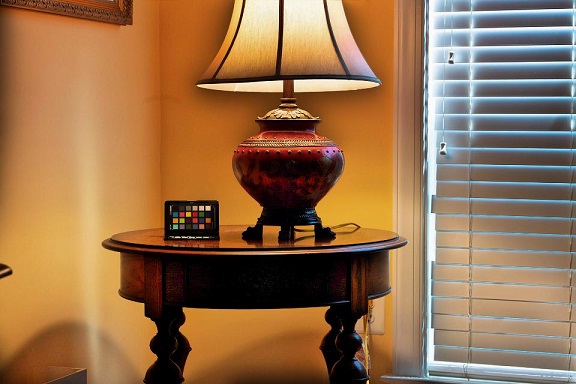}
		\caption{WVM~\cite{fu2016weighted}}
	\end{subfigure}
	\begin{subfigure}[c]{0.23\textwidth}
		\centering
		\includegraphics[width=1.65in]{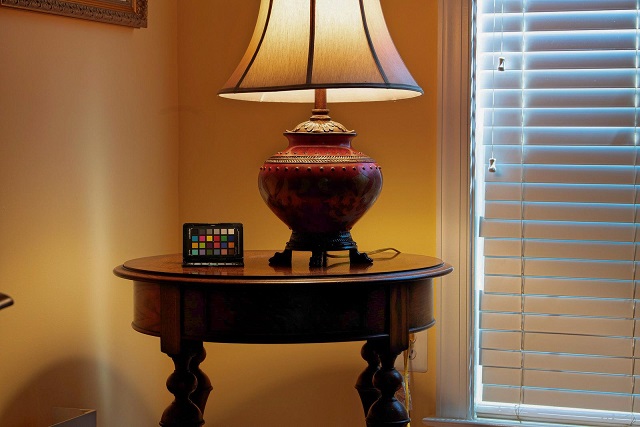}
		\caption{JieP~\cite{cai2017joint}}
	\end{subfigure}  \\ \vspace{2mm}
	\begin{subfigure}[c]{0.23\textwidth}
		\centering
		\includegraphics[width=1.65in]{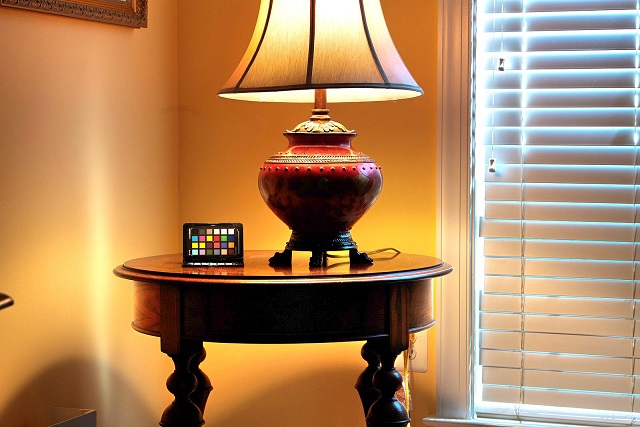}
		\caption{LIME~\cite{guo2017lime}}
	\end{subfigure}
	\begin{subfigure}[c]{0.23\textwidth}
		\centering
		\includegraphics[width=1.65in]{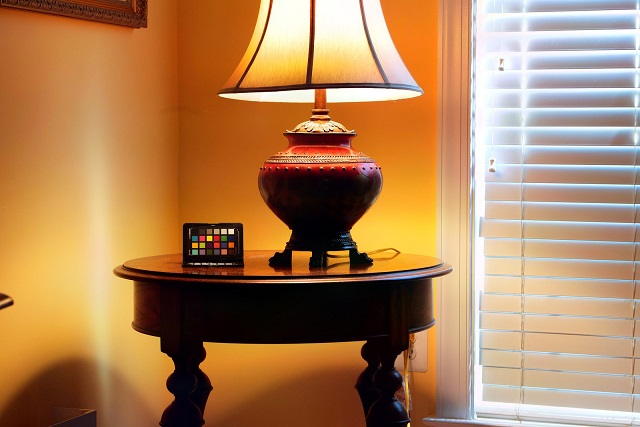}
		\caption{HDRNet~\cite{gharbi2017deep}}
	\end{subfigure}
	\begin{subfigure}[c]{0.23\textwidth}
		\centering
		\includegraphics[width=1.65in]{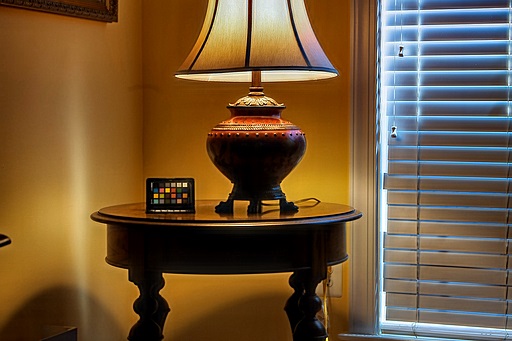}
		\caption{DPE~\cite{chen2018deep}}
	\end{subfigure}
	\begin{subfigure}[c]{0.23\textwidth}
		\centering
		\includegraphics[width=1.65in]{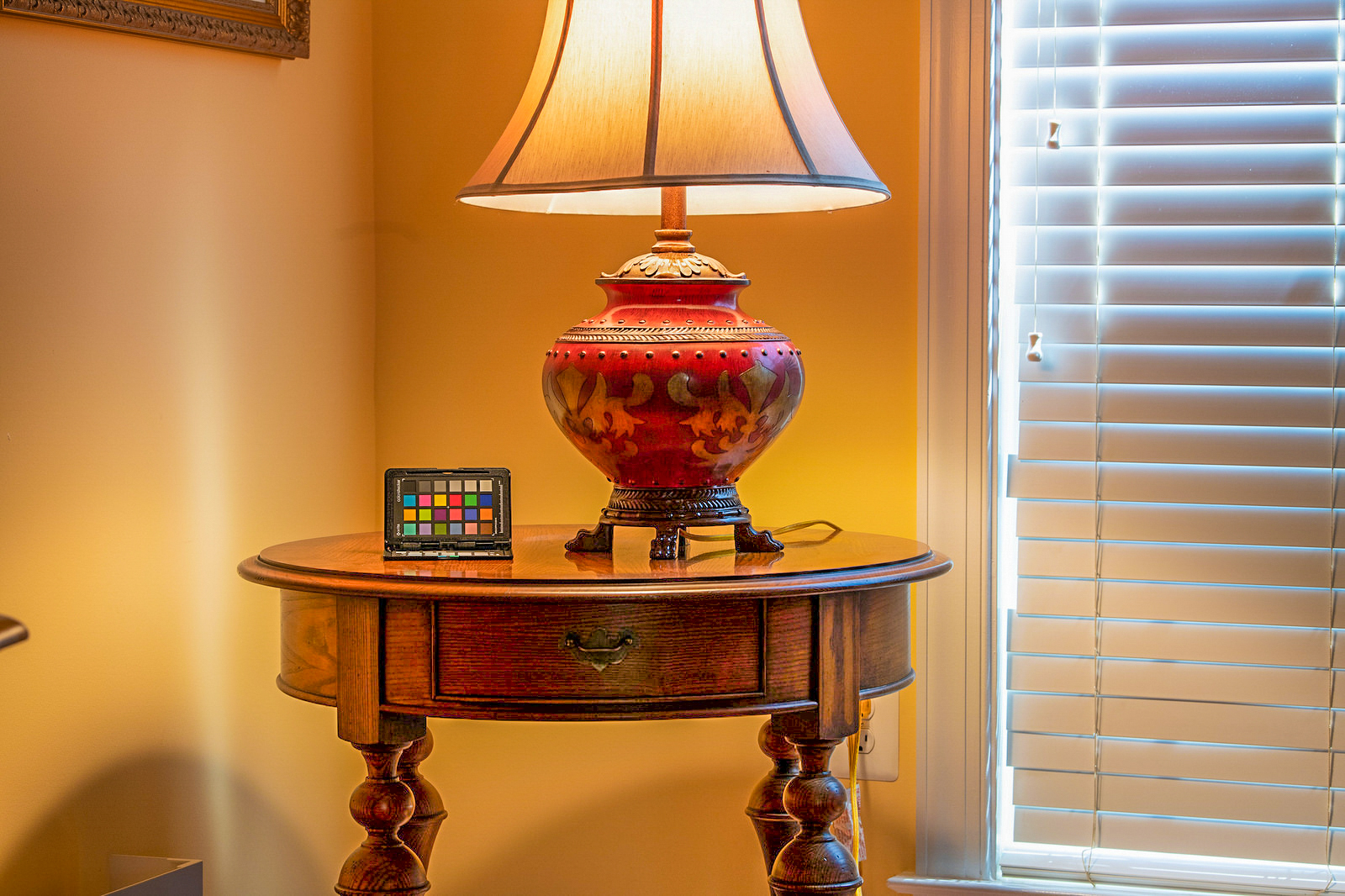}
		\caption{Ours}
	\end{subfigure}
	\caption{Comparison between our method and the state-of-the-art methods on enhancing a challenging underexposed photo. }
	\label{fig:teaser_comp_method}
    \vspace{-2mm}
\end{figure*}

There have been various underexposed photo enhancement algorithms in the research community. Early approaches work by performing contrast enhancement \cite{xu2013generalized,wang2015guided}. Many subsequent approaches~\cite{fu2015probabilistic,gao2017naturalness,cai2017joint,guo2017lime,fu2019towards} rely on the Retinex theory~\cite{land1977retinex}, the camera response function~\cite{ying2017new}, and the inverted images~\cite{dong2011fast,li2015low} to enhance photos. Others learn data-driven photo adjustment by utilizing either traditional machine learning techniques~\cite{bychkovsky2011learning,hwang2012context,yan2014learning}, or deep neural networks~\cite{gharbi2017deep,park2018distort,yu2018deepexposure,chen2018learning,chen2018deep,wang2019underexposed,zhang2019kindling}. However, as shown in Fig.~\ref{fig:teaser_comp_method}, these methods still have respective limitations, \eg, the unclear details, local overexposure and color distortion, making they fail to produce visually pleasing results.

To address the limitations of previous methods, we present a novel method for enhancing underexposed photos. Our method is built upon the observation that the main reason why existing methods produce visually unpleasing results is because they may break the perceptually consistency of visual information between the underexposed input and its enhanced output. For instance, an enhanced image with loss of detail issue is unsatisfactory, since they break the edge consistency with the input image. Based on this observation, we propose perceptually bidirectional similarity (PBS) for explicitly enforcing the perceptual consistency, and formulate underexposed photo enhancement as PBS-constrained illumination estimation by defining PBS as constraints on illumination, which allows us to recover high-quality results from the acquired illumination. Besides, an illumination-estimation-based video enhancement framework is described to handle underexposed videos, where we sample keyframes for illumination estimation and then propagate the illuminations of keyframes to other video frames in a temporally coherent fashion via a Bayesian formulation.

In summary, this paper presents:
\begin{itemize}
	\item First, we propose PBS, a simple yet effective criterion for explicitly describing how to ensure the perceptual consistency during underexposed photo enhancement.
	\item Second, we design PBS-constrained illumination estimation for enhancing underexposed photos in a way that avoids the artifacts encountered by previous methods.
	\item Third, we adopt the proposed illumination estimation and introduce an underexposed video enhancement framework, which produces very competitive video enhancement results compared to existing methods.
	\item
	Fourth, we evaluate the performance of our method in enhancing underexposed photos on six datasets and compare it with various state-of-the-art methods. Results show that our method outperforms previous methods.
\end{itemize}

A preliminary version of this work appeared in~\cite{zhang2018high}. In this paper, we have extended the earlier conference version in four aspects. First, we present an effective video enhancement framework based on the proposed PBS-constrained illumination estimation. In particular, a probabilistic illumination propagation approach is introduced to obtain temporally coherent illumination sequence for an input video from illuminations of sampled keyframes. Second, we introduce an efficient implementation for our illumination estimation. Third, we provide deeper analysis to our method, including the relationship to color constancy and the potential in correcting overexposed images, etc. Fourth, we have conducted extensive experiments to evaluate the advantage of our method, including further comparisons with more recent learning-based methods and evaluations on additional datasets.

\section{Related Work} \label{sec:related_work}
\noindent \textbf{Histogram-based methods.} One of the most widely-adopted image enhancement techniques is histogram equalization (HE), which increases image contrast by finding a transformation function that evens out the intensity histogram. However, it tends to cause loss of contrast for regions with high frequencies. To improve the result, Zuiderveld~\etal~\cite{zuiderveld1994contrast} presented the contrast limited adaptive histogram equalization (CLAHE) by setting a limit on the derivative of the slope of the transformation function. This method is quite effective in contrast enhancement, but may induce ghosting artifacts. Although there are many subsequent HE-based variants~\cite{celik2011contextual,xu2013generalized}, they may also produce unsatisfactory results.

\begin{figure*}
	\centering
	\begin{subfigure}[c]{0.23\textwidth}
		\includegraphics[width=1.65in]{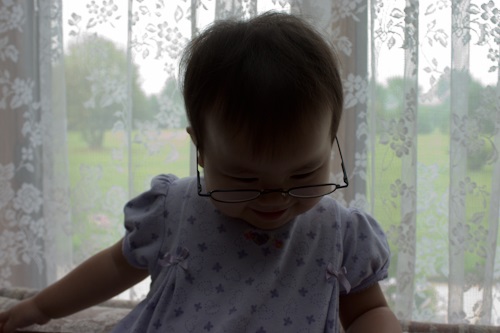}
		\caption{Input}
	\end{subfigure}
	\begin{subfigure}[c]{0.23\textwidth}
		\includegraphics[width=1.65in]{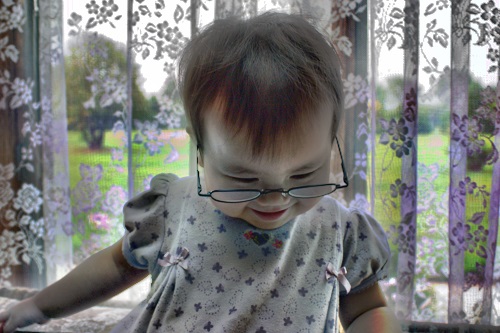}
		\caption{CLAHE~\cite{zuiderveld1994contrast}}
	\end{subfigure}
	\begin{subfigure}[c]{0.23\textwidth}
		\includegraphics[width=1.65in]{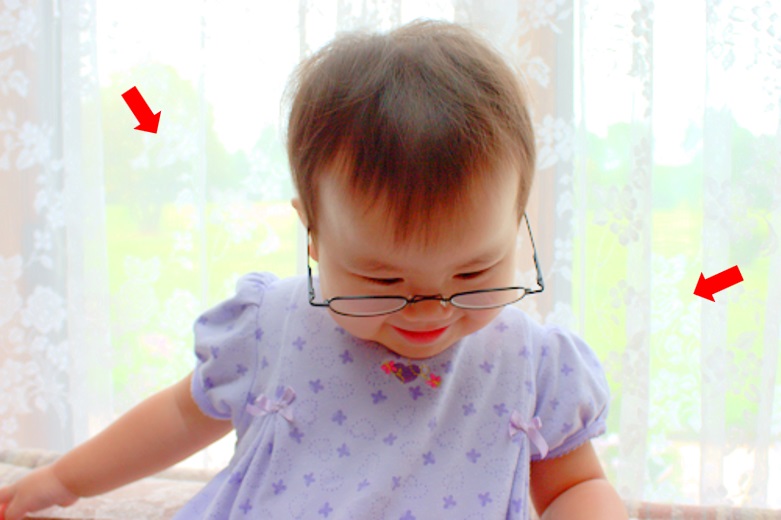}
		\caption{Bennett and McMillan~\cite{bennett2005video}}
	\end{subfigure}
	\begin{subfigure}[c]{0.23\textwidth}
		\includegraphics[width=1.65in]{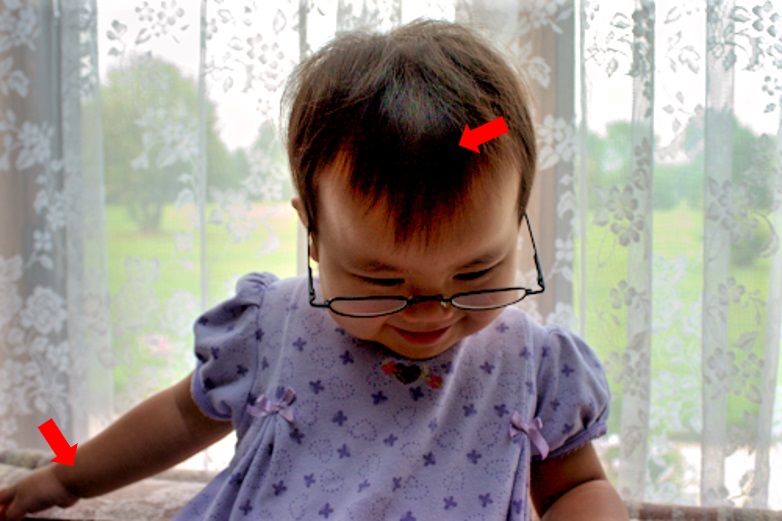}
		\caption{Yuan and Sun~\cite{yuan2012automatic}}
	\end{subfigure}   \\ \vspace{2mm}
	
	\begin{subfigure}[c]{0.23\textwidth}
		\includegraphics[width=1.65in]{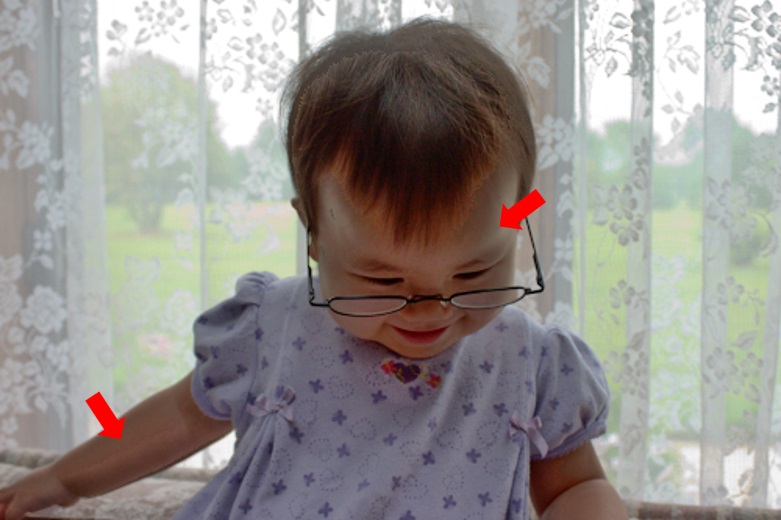}
		\caption{NPE~\cite{wang2013naturalness}}
	\end{subfigure}
	\begin{subfigure}[c]{0.23\textwidth}
		\includegraphics[width=1.65in]{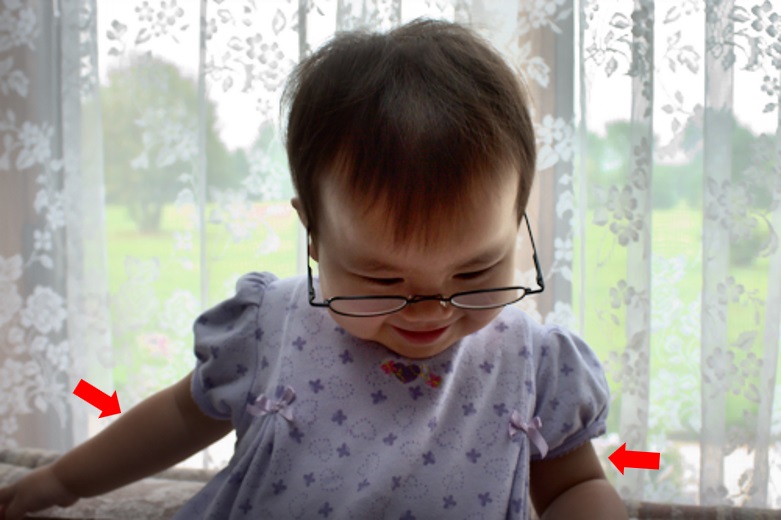}
		\caption{WVM~\cite{fu2016weighted}}
	\end{subfigure}
	\begin{subfigure}[c]{0.23\textwidth}
		\includegraphics[width=1.65in]{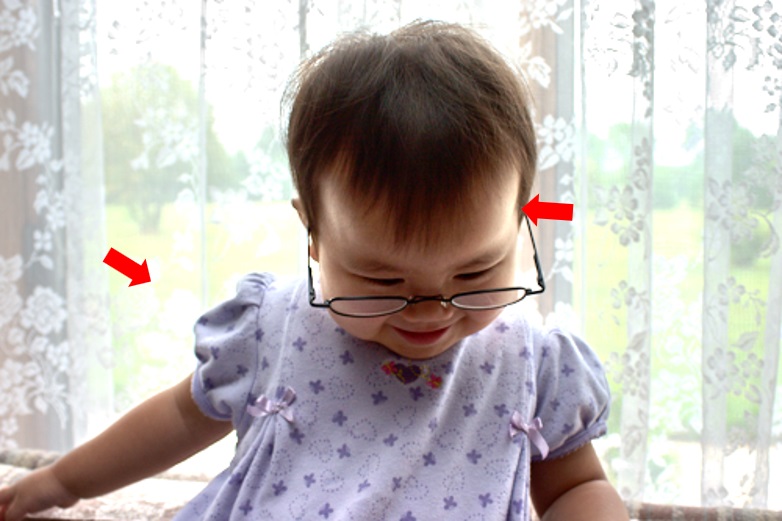}
		\caption{LIME~\cite{guo2017lime}}
	\end{subfigure}
	\begin{subfigure}[c]{0.23\textwidth}
		\includegraphics[width=1.65in]{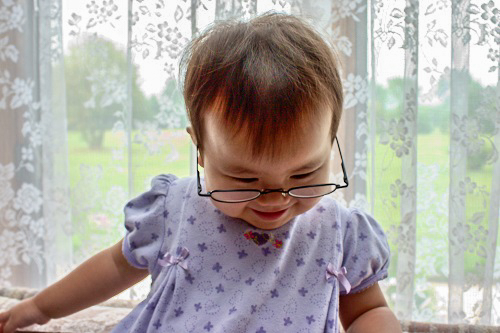}
		\caption{Ours}
	\end{subfigure}
	\caption{Common issues encountered by existing underexposed photo enhancement methods. Photo from Bychkovsky~\etal~\cite{bychkovsky2011learning}.}
	\label{fig:issues}
	\vspace{-2mm}
\end{figure*}

\vspace{0.3em}
\noindent \textbf{Sigmoid-mapping-based methods.} Mapping pixel intensities with sigmoid functions is another way to enhance underexposed images. As globally applying sigmoid mapping may generate visually distorted results, existing methods usually perform local intensity mapping. For instance,
Bennett and McMillan~\cite{bennett2005video} decomposed the input image into a base layer and a detail layer, and then applied different mappings for the two layers to preserve the image details. Yuan and Sun~\cite{yuan2012automatic} segmented the input image into subregions and computed luminance-aware detail-preserving mapping for each subregion. Zhang~\etal~\cite{zhang2016underexposed} created multiple tone-mapped versions for the input image and fused them into a well-exposed image. Since finding locally optimal sigmoid mappings and ensuring globally smooth transitions are difficult, these methods may not work well for images with uneven exposure.

\vspace{0.3em}
\noindent \textbf{Retinex-based methods.} This kind of method is built upon the assumption that an underexposed image is the pixel-wise product of the expected enhanced image and a single-channel illumination map. In this way, image enhancement can be reduced to an illumination estimation problem. Jobson~\etal~\cite{jobson1997multiscale} made an early attempt to this problem, but their results often look unnatural. Although subsequent methods significantly improve the results~\cite{wang2013naturalness,fu2015probabilistic,fu2016fusion, fu2016weighted,cai2017joint,guo2017lime,wei2018deep}, they may also induce visual artifacts such as loss of details, color distortion and uneven exposure. Our method also belongs to this category, which extends upon the previous work~\cite{zhang2018high} in four different ways as mentioned in the introduction, and is able to robustly generate visually pleasing results free of the visual artifacts encountered by previous methods.

\vspace{0.3em}
\noindent \textbf{Learning-based methods.} An increasing amount of efforts focus on investigating learning-based methods since the pioneering work of Bychkovsky~\etal~\cite{bychkovsky2011learning}, which provides a dataset consisting of image pairs for tone adjustment. Yan~\etal~\cite{yan2014learning} achieved automatic color enhancement by tackling a learning-to-rank problem, while Yan~\etal~\cite{yan2016automatic} enabled semantic-aware image enhancement by leveraging scene semantics. Gharbi~\etal~\cite{gharbi2017deep} proposed bilateral learning to enable real-time image enhancement, while Chen~\etal~\cite{chen2018deep} designed an unpaired learning model for enhancement based on a two-way generative adversarial networks (GANs). Yang~\etal~\cite{yang2018image} corrected LDR images by using a deep reciprocating HDR transformation. Cai~\etal~\cite{cai2018learning} learned a contrast enhancer from multi-exposure images. Deep encoder-decoder network is also utilized to enhance low-light images~\cite{lore2017llnet,ren2019low}. More recently, Jiang~\etal~\cite{jiang2019enlightengan} introduced the EnlightenGAN for low-light enhancement, while two other recent methods work by performing deep Retinex decomposition~\cite{wang2019underexposed,zhang2019kindling}. However, these methods may not work well on images that are significantly different with the training images.

\section{Underexposed Photo Enhancement} \label{sec:photo_enhancement}
This section presents our underexposed photo enhancement approach. We first summarize the background knowledge on Retinex-based image enhancement and illustrate how to cast photo enhancement as an illumination estimation problem. Then, we introduce PBS and analyze how we define it as constraints on illumination. Next, we formulate PBS-constrained illumination estimation for enhancing underexposed photos while avoiding the common visual artifacts, and provide in-depth model analysis. Finally, we describe an efficient implementation for the illumination estimation.

\subsection{Background on Retinex-based Image Enhancement} \label{sec:method_background}
Retinex-based image enhancement \cite{fu2016weighted,guo2017lime} assumes that an underexposed image $I$ (normalized to [0,1]) is the pixel-wise product of the desired enhanced image $R$ and a single-channel illumination map $S$, which is expressed as
\begin{equation} \label{equ:retinex}
	I = S \times R,
\end{equation}
where $\times$ denotes pixel-wise multiplication. With the above assumption, image enhancement can be reduced to an illumination estimation problem, since the enhanced image can be recovered by $R = I / S$ as long as $S$ is known.

\begin{figure*}
	\centering
	\begin{subfigure}[c]{0.184\textwidth}
		\centering
		\includegraphics[width=1.32in]{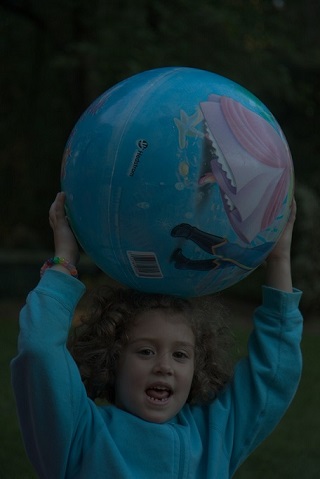}
		\caption{Input}
	\end{subfigure}
	\begin{subfigure}[c]{0.184\textwidth}
		\centering
		\includegraphics[width=1.32in]{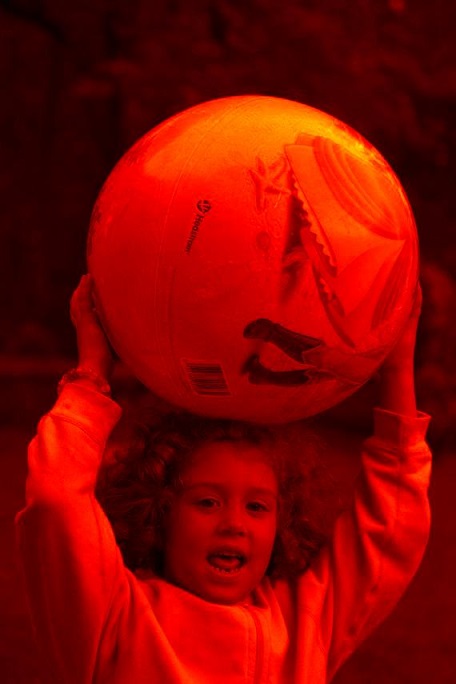}
		\caption{Initial illumination $S'$}
	\end{subfigure}
	\begin{subfigure}[c]{0.184\textwidth}
		\centering
		\includegraphics[width=1.32in]{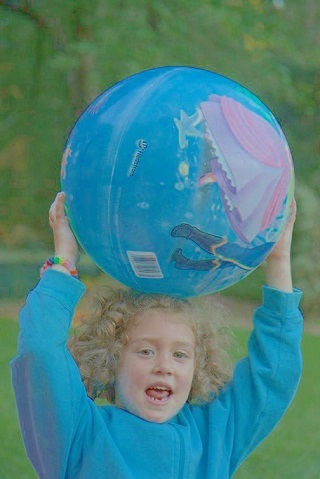}
		\caption{Result from $S'$}
	\end{subfigure}
	\begin{subfigure}[c]{0.184\textwidth}
		\centering
		\includegraphics[width=1.32in]{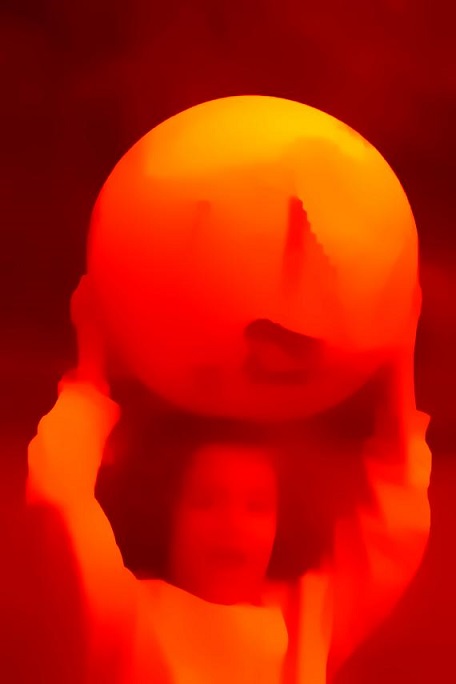}
		\caption{Refined illumination $S$}
	\end{subfigure}
	\begin{subfigure}[c]{0.184\textwidth}
		\centering
		\includegraphics[width=1.32in]{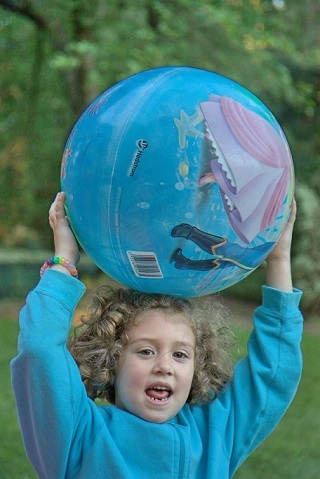}
		\caption{Result from $S$}
	\end{subfigure}
	\caption{An example underexposed photo enhanced by the proposed PBS-constrained illumination estimation. The single channel illuminations (b) and (d) are shown in hot colormap. Source image from Bychkovsky~\etal~\cite{bychkovsky2011learning}.}	
	\label{fig:lum_estimation}
	\vspace{-2mm}
\end{figure*}

\subsection{Perceptually Bidirectional Similarity (PBS)} \label{sec:method_pbs}
We first analyze the common issues encountered by existing methods, which inspire the proposal of PBS. As shown in Fig.~\ref{fig:issues}(b)-(g), color distortion, uneven exposure and loss of detail are the three main issues. CLAHE~\cite{zuiderveld1994contrast} and NPE~\cite{wang2013naturalness} distort the skin color and mistakenly make the girl's face and arms gray, giving rise to color family mismatch between the input image and the enhanced outputs. Yuan and Sun~\cite{yuan2012automatic} and WVM~\cite{fu2016weighted} induce exposure inconsistency around the arms and the body, while these regions have consistent exposure in the input image. Bennett and McMillan~\cite{bennett2005video} and LIME~\cite{guo2017lime} overexpose the background and lead to loss of detail.

From the above analysis, we have come to an important observation --- that is, the reason why existing methods fail to produce visually pleasing results is because they break the perceptual consistency of color, detail and local exposure distribution between the input image and the enhanced output. In other words, this observation suggests that a good enhanced image should not only improve the detail visibility of the underexposed regions, but also satisfy two properties: 1) it should contain all the visual information (can be enhanced versions) in the input image; 2) it should not introduce new visual information that does not exist in the input image. Aware of this, we propose perceptually bidirectional similarity (PBS), which more specifically characterizes the aforementioned two requirements for the enhanced image $R$ of an underexposed image $I$: 1) colors and details in $I$ should all exist in $R$ as properly enhanced versions ($\geq 1$), and regions in $I$ with consistent exposure should also have consistent exposure in $R$; 2) $R$ should not contain distorted colors, additional details and exposure inconsistencies that originally do not exist in $I$.

\subsection{PBS as Constraints on Illumination}
To utilize PBS, we define it as three constraints on illumination $S$, which help ensure the bidirectional perceptual consistency of color, detail and exposure distribution between the input image $I$ and the enhanced image $R$, respectively.

\vspace{0.3em}
\noindent \textbf{Color consistency.} To preserve color consistency, we enforce each pixel's color in $R$ and $I$ are in the same color family by imposing a range constraint on $S$. Since $R = I / S$ and $I$ is normalized to [0,1], small (large) $S$ yields $R$ with high (low) RGB values. Intuitively, color inconsistency may appear in terms of mismatched colors in $R$ derived from naive color truncation, when $S$ is too small to guarantee that each RGB color channel in $R$ remains in the color gamut [0,1]. Hence, we bound $S$ to be no less than a value that can enlarge the maximum RGB color channel of each pixel in $I$ to the upper bound 1 through $R = I / S$, which is expressed as
\begin{equation} \label{equ:color_range}
	\max I^c_p = \Gamma(S^{\min }_p),~~\forall c \in \{ r,g,b\},
\end{equation}
where $I^c_p$ is a color channel at pixel $p$. $\Gamma(\alpha) = \alpha^\gamma$ is the Gamma function with $\gamma \in (0,1)$, which is an optional operation for further illumination adjustment. From Eq.~\ref{equ:color_range}, we can easily obtain $S^{\min }_p = (\max I^c_p)^{1 / \gamma}$. To avoid mistakenly darken the input underexposed image, we set the upper bound of $S$ to 1, in which case the input will be directly taken as the output. Overall, for each pixel $p$, the color consistency constraint can be defined as $S^{\min }_p \leq S_p \leq 1$.

\vspace{0.3em}
\noindent \textbf{Detail consistency.} We formulate the detail consistency described by PBS from a perspective of edge consistency as follows: 1) If $I$ is smooth at pixel $p$, then $R$ should also be smooth at $p$; 2) If $I$ has an edge at pixel $p$, then $R$ should have a stronger, or at least equivalent edge at $p$. By associating edge with gradient and directional derivative, the above two cases can be characterized as the following constraint:
\begin{equation} \label{equ:pbs_details}
	\left\{
	{\begin{array}{lc}
			{\nabla R_p  = 0,} & {\left|\nabla I_p \right| \leq \tau } \\
			{{\partial_d R_p}/{\partial_d I_p} \geq 1,} & {\left|\nabla I_p  \right| > \tau}  	
	\end{array}} \right.
\end{equation}
where $\nabla$ denotes the gradient operator. $\partial_{d \in \{x, y\}}$ is the first order derivative along the horizontal ($x$) or vertical ($y$) direction. $\tau$ is a small constant (typically 1e-5) for determining whether there is an edge at a pixel.  Note Eq.~\ref{equ:pbs_details} can also be expressed as formulation about $S$ by replacing $R$ with $I / S$.

\begin{figure*}
	\centering
	\begin{subfigure}[c]{0.184\textwidth}
		\centering
		\includegraphics[width=1.32in]{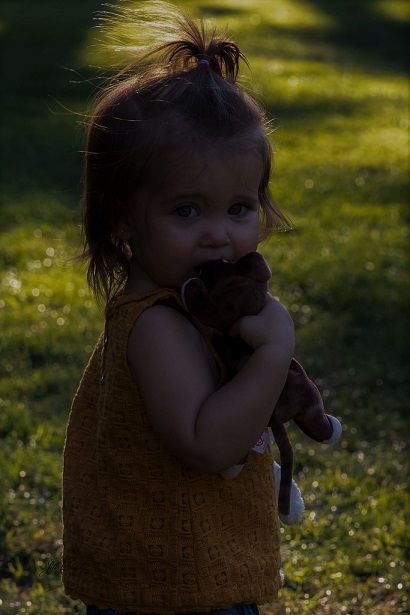}
		\caption{}
	\end{subfigure}
	\begin{subfigure}[c]{0.184\textwidth}
		\centering
		\includegraphics[width=1.32in]{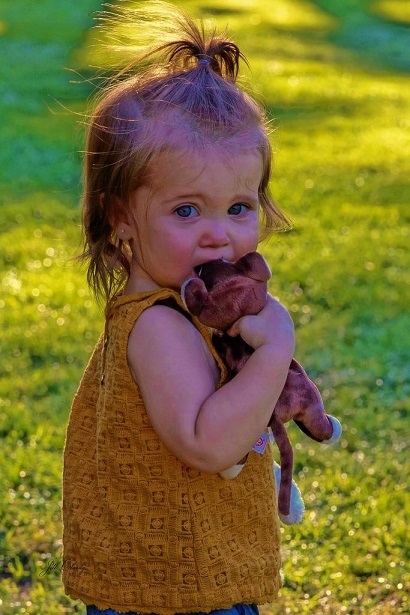}
		\caption{}
	\end{subfigure}
	\begin{subfigure}[c]{0.184\textwidth}
		\centering
		\includegraphics[width=1.32in]{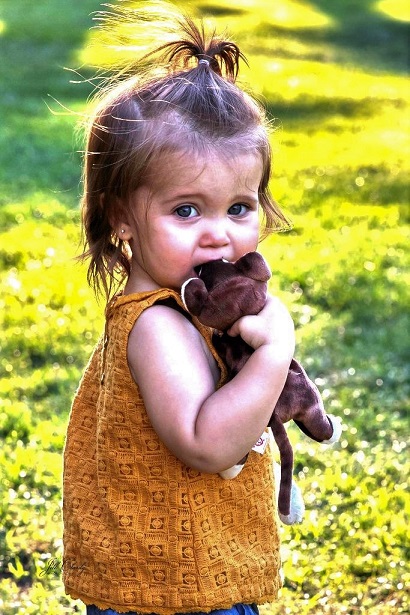}
		\caption{}
	\end{subfigure}
	\begin{subfigure}[c]{0.184\textwidth}
		\centering
		\includegraphics[width=1.32in]{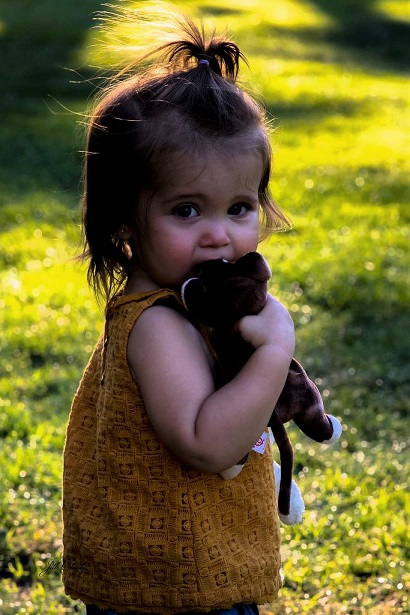}
		\caption{}
	\end{subfigure}
	\begin{subfigure}[c]{0.184\textwidth}
		\centering
		\includegraphics[width=1.32in]{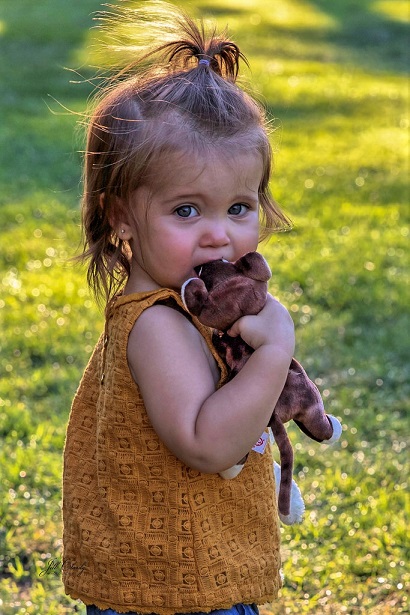}
		\caption{}
	\end{subfigure}
	\caption{Effectiveness of each PBS constraint. (a) Input image. (b)-(d) are enhanced images without color, detail and exposure distribution consistency constraints, respectively. (e) Result with all the three PBS constraints.}
	\label{fig:pbs_validation}
	\vspace{-2mm}
\end{figure*}

\vspace{0.3em}
\noindent \textbf{Exposure distribution consistency.} According to Eq.~\ref{equ:retinex}, the key to preserving the exposure distribution consistency is to ensure that $S$ is locally smooth for regions with similar brightness in the input. To this end, we alternatively adopt the relative total variation (RTV) measure \cite{xu2012structure} as the smoothness regularizer for obtaining piecewise smooth illumination, while maintaining prominent illumination discontinuities across regions. Adopting this regularizer can also help enhance image contrast, because when adjacent pixels $p$ and $q$ have similar illumination values ($S_p \approx S_q$), their contrast in the enhanced image $R$ can be estimated as $|R_p-R_q|\approx|I_p-I_q| / S_p$, which will be enhanced, since $S \leq 1$. Note, other edge-aware smoothness regularizers \cite{farbman2008edge,xu2011image} can also work with our approach. Formally, the RTV measure is defined as
\begin{equation} \label{equ:rtv_all}
	RTV(S_p) = \mathcal{H}(S_p) + \mathcal{V}(S_p),
\end{equation}
where $\mathcal{H}(S_p)$ and $\mathcal{V}(S_p)$ denote the $x$- and $y$-direction RTV measure, respectively. Specifically, the $x$-direction measure $\mathcal{H}(S_p)$ is written as
\begin{equation}
	\mathcal{H}(S_p) = \sum\limits_{q\in \mathcal{N}_p}u^x_qw^x_q(\partial_x S_q)^2,
\end{equation}
where $\mathcal{N}_p$ denotes a $15 \times 15$ window centered at pixel $p$. $u^x_q = G_{\sigma}*(|G_{\sigma} * \partial_xS_q| + \epsilon)^{-1}$ and $w^x_q = (|\partial_xS_q| + \epsilon)^{-1}$, where $G_{\sigma}$ denotes a Gaussian kernel with standard deviation $\sigma = 3$, $*$ is the convolution operator, and $\epsilon=1e-3$ is used for preventing division by zero. $\mathcal{V}(S_p)$ is defined similarly.

\subsection{PBS-constrained Illumination Estimation} \label{sec:method_illumination_estimation}
This section illustrates how we formulate underexposed photo enhancement as PBS-constrained illumination estimation. We first introduce how to obtain an initial illumination for an input image. Then, we adopt the PBS constraints and design an optimization framework for refining the initial illumination, so that we can obtain the illumination that is able to recover PBS-satisfied enhanced image.

Intuitively, the brightness of different areas in an image roughly reflect the magnitude of illumination. Hence, inspired by \cite{land1977retinex}, we obtain the initial illumination $S'$ by treating the maximum values among the RGB color channels of the input image $I$ as the illumination values, which is expressed as
\begin{equation} \label{equ:init_lum}
	S'_p = \max I^c_p,~~\forall c \in \{r, g, b\}.
\end{equation}
As analyzed by \cite{guo2017lime}, by this means, the initial illumination can better model the global illumination distribution, and also ensures that the enhanced image $R$ will be less saturated.

Although the initial illumination roughly depicts the overall illumination distribution, it typically contains richer details and textures that are not led by illumination discontinuities, making enhanced image directly recovered from it visually unrealistic, as shown in Fig.~\ref{fig:lum_estimation}(c). Hence, we propose to estimate a refined illumination $S$ that satisfies the PBS constraints on illumination. To this end, we formulate the following objective function for estimating the desired illumination $S$:
\begin{equation} \label{equ:obj_func}
	\begin{split}
		&\mathop{\arg \min}\limits_S \sum\limits_p {(S_p  - S'_p )^2} + \lambda \Big(\mathcal{H}(S_p) + \mathcal{V}(S_p)\Big), ~~s.t.~~\\
		&S^{\min }_p \leq S_p \leq 1, \left\{
		{\begin{array}{lc}
				{\nabla (I_p / S_p)  = 0,} & {\left|\nabla I_p \right| \leq \tau } \\
				{{\partial_d (I_p / S_p)}/{\partial_d I_p} \geq 1,} & {\left|\nabla I_p  \right| > \tau}  	
		\end{array}} \right.
	\end{split}
\end{equation}
where $\lambda$ is the balancing weight. The first term $(S_p - S'_p)^2$ forces the target illumination to be close to the initial illumination in structure, while the second term and the other two constraints are the PBS constraints. The objective function in Eq.~\ref{equ:obj_func} can be solved by introducing auxiliary variables to divide the intractable problem into several tractable subproblems (see \cite{zhang2018high} for details). With the refined illumination $S$, the final enhanced image is recovered by $R = I/S^{\gamma}$. Fig.~\ref{fig:lum_estimation} shows an example image enhanced by the proposed PBS-constrained illumination estimation. We can see that the refined illumination removes the redundant texture details in the initial illumination and yields more appealing enhancement result.

\begin{figure*}
	\centering
	\captionsetup[subfigure]{labelformat=empty}
	\begin{subfigure}[c]{0.23\textwidth}
		\centering
		\includegraphics[width=1.65in]{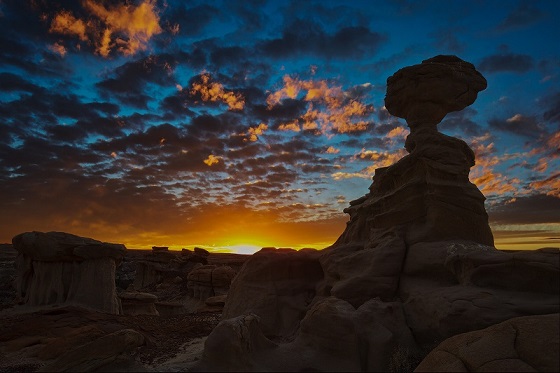}
		\caption{Image 1}
	\end{subfigure}
	\begin{subfigure}[c]{0.23\textwidth}
		\centering
		\includegraphics[width=1.65in]{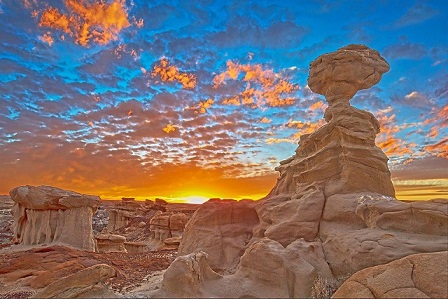}
		\caption{\bm{$\lambda = 0.1$}, $\gamma=0.6$}
	\end{subfigure}
	\begin{subfigure}[c]{0.23\textwidth}
		\centering
		\includegraphics[width=1.65in]{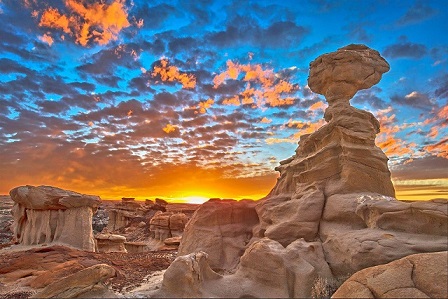}
		\caption{\bm{$\lambda = 0.8$}, $\gamma=0.6$}
	\end{subfigure}
	\begin{subfigure}[c]{0.23\textwidth}
		\centering
		\includegraphics[width=1.65in]{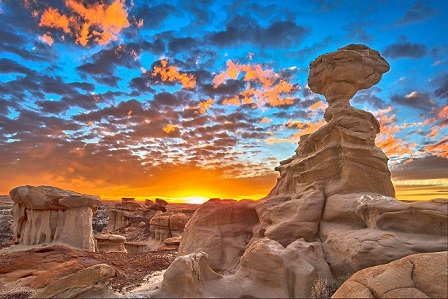}
		\caption{\bm{$\lambda = 2.0$}, $\gamma=0.6$}
	\end{subfigure} \\ \vspace{2mm}
	
	\begin{subfigure}[c]{0.23\textwidth}
		\centering
		\includegraphics[width=1.65in]{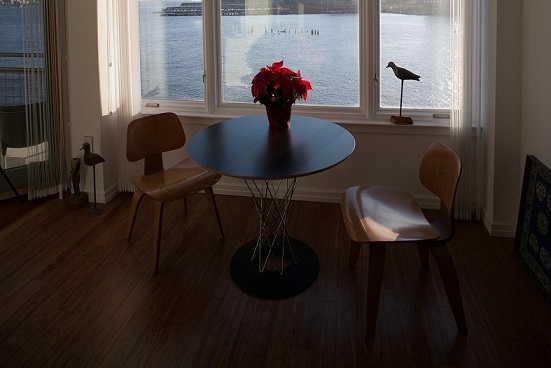}
		\caption{Image 2}
	\end{subfigure}
	\begin{subfigure}[c]{0.23\textwidth}
		\centering
		\includegraphics[width=1.65in]{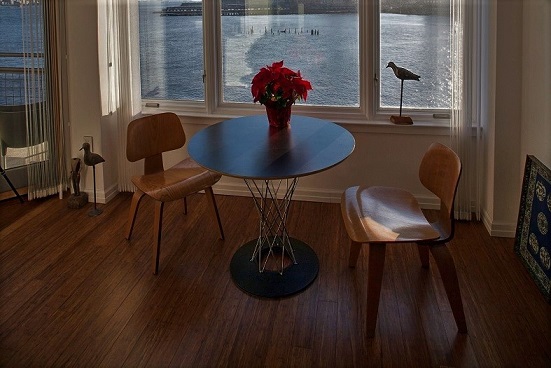}
		\caption{$\lambda = 0.8$, \bm{$\gamma=0.3$}}
	\end{subfigure}
	\begin{subfigure}[c]{0.23\textwidth}
		\centering
		\includegraphics[width=1.65in]{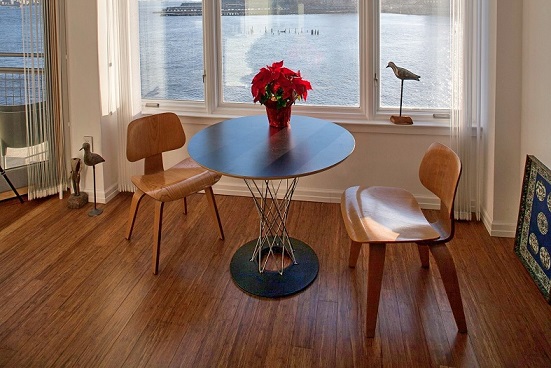}
		\caption{$\lambda = 0.8$, \bm{$\gamma=0.6$}}
	\end{subfigure}
	\begin{subfigure}[c]{0.23\textwidth}
		\centering
		\includegraphics[width=1.65in]{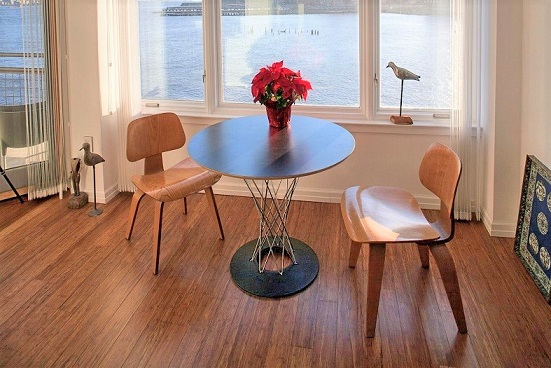}
		\caption{$\lambda = 0.8$, \bm{$\gamma=1.0$}}
	\end{subfigure}
	\caption{Effect of varying $\lambda$ and $\gamma$. The 1st and 2nd rows show how $\lambda$ and $\gamma$ affect the enhanced images, respectively.}
	\label{fig:parameters}
	\vspace{-2mm}
\end{figure*}

\subsection{Model Analysis} \label{sec:method_implementation}
\noindent \textbf{Effectiveness of each PBS constraint.} Fig.~\ref{fig:pbs_validation} validates the effectiveness of each PBS constraint. We can see that the skin color is obviously distorted when we remove the color consistency constraint (see Fig.~\ref{fig:pbs_validation}(b)), while removing the detail consistency constraint makes the grass as well as the face and arm overexposed (see Fig.~\ref{fig:pbs_validation}(c)). Without the exposure distribution consistency constraint, the enhanced image shows unpleasing exposure inconsistency around the body (see Fig.~\ref{fig:pbs_validation}(d)), while these regions have similar exposure level in the input image. Last, by combining all the three PBS constraints, we obtain a visually pleasing result with clear details, vivid color, distinct contrast and consistent exposure distribution, as shown in Fig.~\ref{fig:pbs_validation}(e).

\vspace{0.3em}
\noindent \textbf{Parameter setting.} The key parameter of our approach is $\lambda$, which determines the smoothness level of the estimated illumination. In general, we set large $\lambda$ for highly textured images. $\gamma$ is another parameter that affects the result quality. In all our experiments, we empirically set $\lambda = 0.8$ and $\gamma = 0.6$, which are able to produce reasonably good results for our test images. Fig.~\ref{fig:parameters} evaluates the effect of varying $\lambda$ and $\gamma$. As shown in the first row, large $\lambda$ produces result with strong local contrast. However, this effect becomes less obvious when $\lambda > 0.8$. As large $\lambda$ typically requires more iterations to converge, we fix $\lambda = 0.8$ as a trade-off. The second row of Fig.~\ref{fig:parameters} shows how $\gamma$ affects the results. We can see that the result without Gamma adjustment (namely $\gamma=1$) is also satisfactory, but too bright to be consistent with the image aesthetic. Decreasing $\gamma$ reduces the overall brightness, but at the cost of lowering the overall visibility. To obtain better visual results, we set $\gamma=0.6$ for our test images.

\begin{figure}
	\centering
	\includegraphics[width=3.2in]{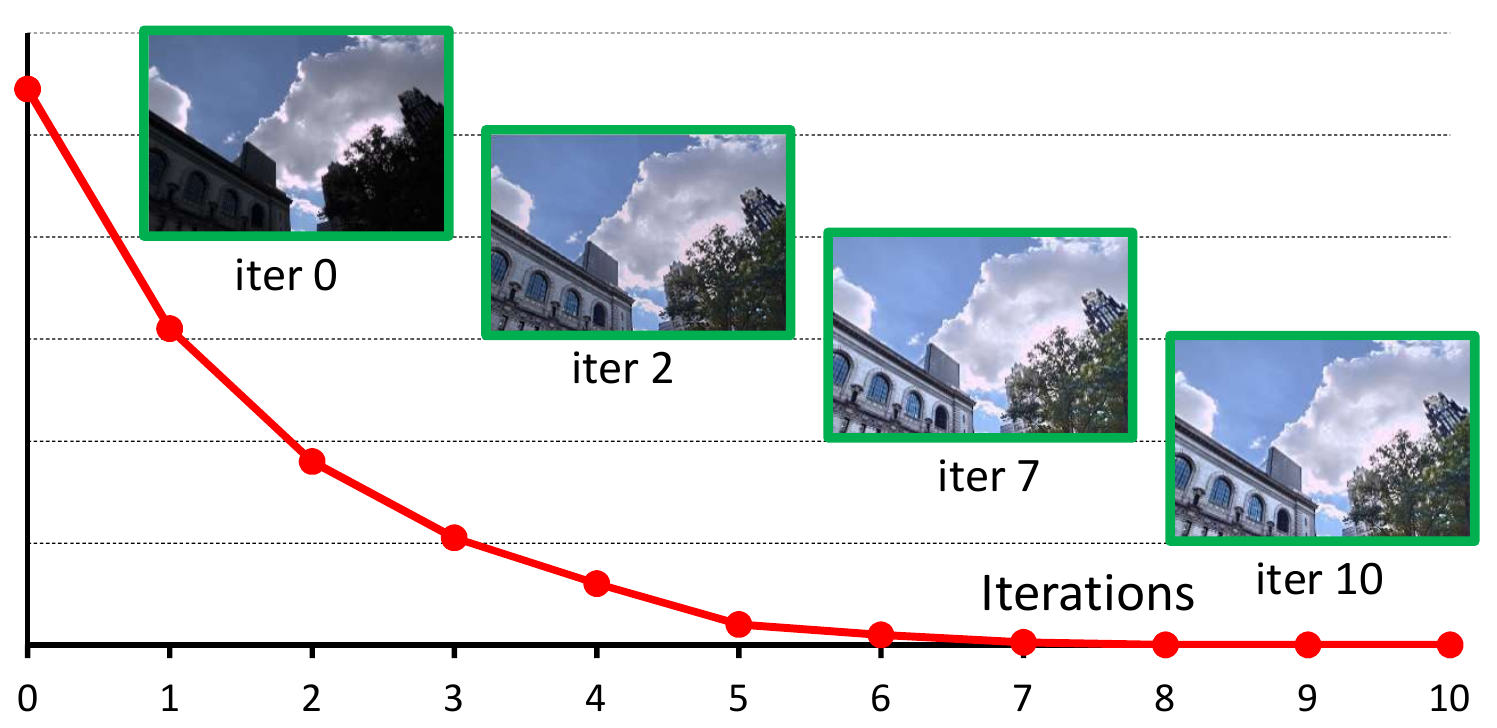}
	\caption{Convergence curve of our PBS-constrained illumination estimation for an example image. The ordinate axis indicates the iterative error of the solutions.}
	\label{fig:convergence}
	\vspace{-2mm}
\end{figure}

\vspace{0.3em}
\noindent \textbf{Convergence analysis.} The PBS-constrained illumination estimation optimization in Eq.~\ref{equ:obj_func} stops iteration when: (i) the difference between two consecutive solutions is less than a small threshold (1e-3), or (ii) the maximum number of iterations (we empirically set it as 20)  is reached. Fig.~\ref{fig:convergence} shows the convergence curve for an example image. As shown, the illumination estimation converges after 7 iterations, and more iterations barely improve the result.

\begin{figure}
	\centering
	\begin{subfigure}[c]{0.147\textwidth}
		\centering
		\includegraphics[width=1.03in]{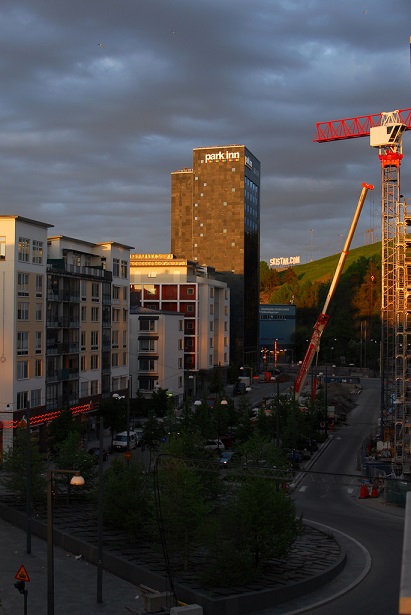} \\ \vspace{1mm}
		\includegraphics[width=1.03in]{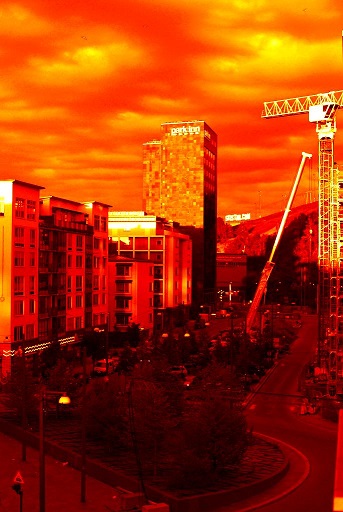}
		\caption{Input}
	\end{subfigure}
	\begin{subfigure}[c]{0.147\textwidth}
		\centering
		\includegraphics[width=1.03in]{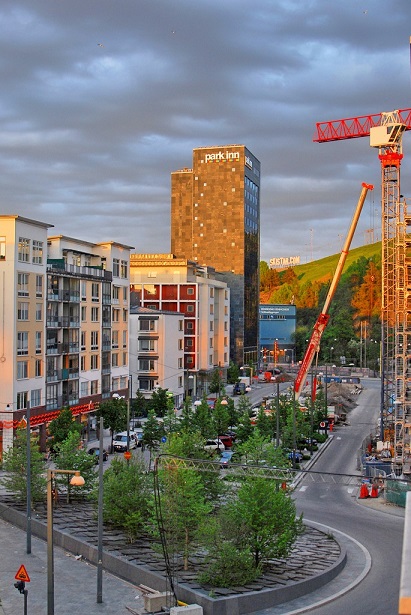} \\ \vspace{1mm}
		\includegraphics[width=1.03in]{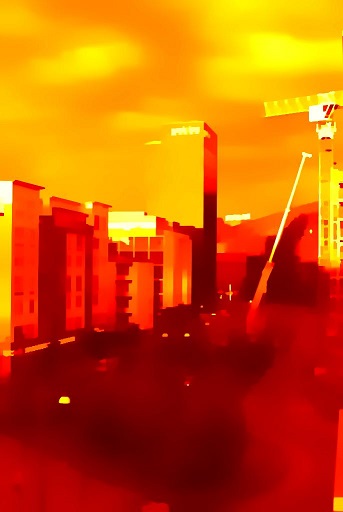}
		\caption{Naive (3 sec)}
	\end{subfigure}
	\begin{subfigure}[c]{0.147\textwidth}
		\centering
		\includegraphics[width=1.03in]{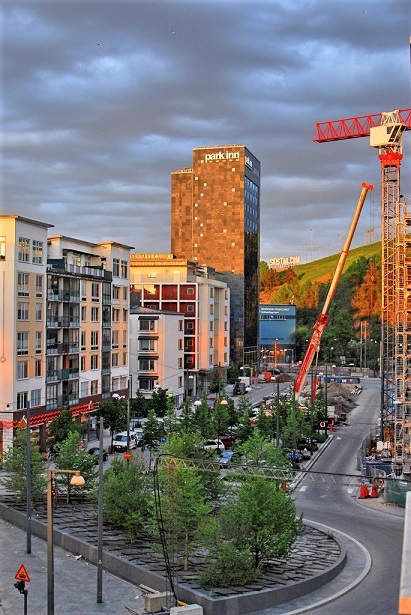} \\ \vspace{1mm}
		\includegraphics[width=1.03in]{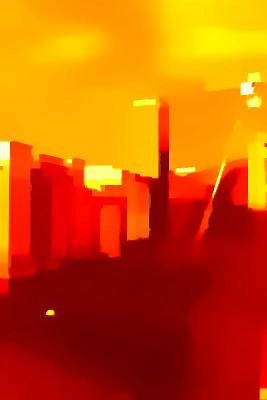}
		\caption{Efficient (0.3 sec)}
	\end{subfigure}
	\caption{Effectiveness of the efficient implementation. The two enhanced images (b) and (c) in the top row are visually indistinguishable, while the efficient implementation takes 0.3 seconds, which is $10 \times$ faster than the naive implementation. The bottom row shows the illuminations.}
	\vspace{-2mm}
	\label{fig:acceleration}
\end{figure}

\begin{figure}
	\centering
	\includegraphics[width=3.4in]{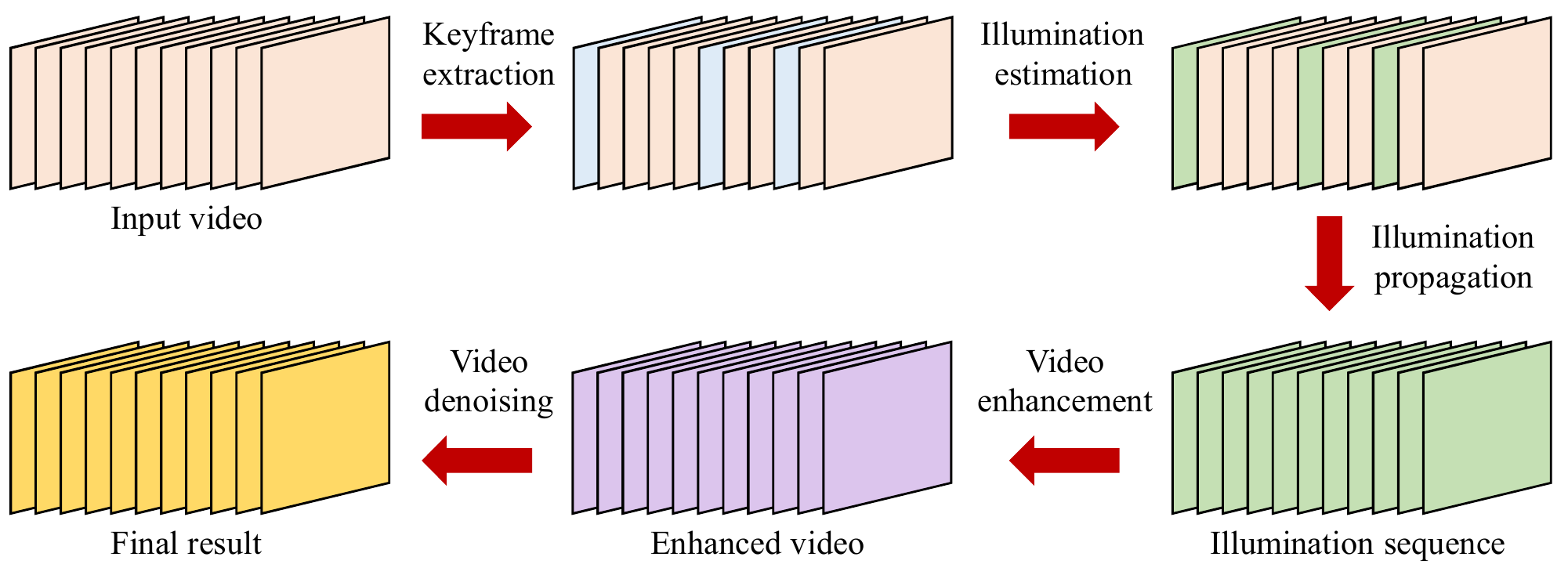}
	\caption{Pipeline of our video enhancement framework. For an input video, we first sample some keyframes that roughly describe the overall illumination changes of the video, and then obtain their illuminations by performing PBS-constrained illumination estimation for each keyframe. Next, we propagate the obtained illuminations to the entire video, and recover the enhanced video based on the acquired illumination sequence. Finally, a video denoising operation is applied to the enhanced video to get the final noise suppressed result.}
	\label{fig:video_pipeline}
    \vspace{-2mm}
\end{figure}

\subsection{Efficient Implementation} \label{sec:method_acceleation}
The PBS-constrained illumination estimation in Eq.~\ref{equ:obj_func} runs practically slow compared with \cite{fu2016fusion,guo2017lime,gharbi2017deep}, because it involves iteratively solving a set of subproblems. To make it more efficient and scalable to high-resolution images, we introduce an efficient computation for it.

Considering that illumination in natural images is generally piece-wise smooth and very suitable for edge-aware sampling, we propose to sample a low-resolution (low-res) input for illumination estimation, and upsample the low-res illumination to full-resolution (full-res) for enhancing the full-res underexposed image. Specifically, we first downsample the input image with its larger dimension (width or height) no more than 400 pixels, and perform illumination estimation on the downsampled low-res input. Then, we apply joint bilateral upsampling (JBU)~\cite{kopf2007joint} to transform the low-res illumination $\bar{S}$ to full-res version $S$ in an edge-aware manner, which is expressed as
\begin{equation} \label{equ:jbu}
	S_p = \frac{1}{\mathcal{Z}_p} \sum_{q_{\downarrow} \in \Omega_{p_{\downarrow}} }\bar{S}_{q_{\downarrow}}f(\left\|p_{\downarrow} - q_{\downarrow}\right\|)g(\left\|S'_p - S'_q\right\|),
\end{equation}
where $S'$ is the initial illumination (full-res) obtained from Eq.~\ref{equ:init_lum}. $p$ and $q$ denote coordinates of pixels in $S$ and $S'$. $p_{\downarrow}$ and $q_{\downarrow}$ denote coordinates of pixels in the low-res solution $\bar{S}$. $f$ and $g$ are spatial and range filter kernels in terms of truncated Gaussian with standard deviation $\sigma_d = 0.5$ and $\sigma_r = 0.1$, respectively. $\Omega$ denotes a $5 \times 5$ window centered at pixel $p_{\downarrow}$. $\mathcal{Z}_p$ is the normalizing factor that sums the filter weight $f(\cdot)g(\cdot)$. Using the above implementation, the runtime for enhancing an 685 $\times$ 1024 image in Fig.~\ref{fig:acceleration}(a) drops from 3 seconds to 0.3 seconds on a PC with Core i5-7400 CPU, while the enhanced image is visually indistinguishable from that of the naive implementation, as shown in Fig.~\ref{fig:acceleration}.

\section{Underexposed Video Enhancement}  \label{sec:video_enhancement}
This section describes how we extend our method to handle underexposed videos. As implementing the PBS-constrained illumination estimation for each video frame tends to cause temporal inconsistencies in the form of jittering artifacts, and naively extending the illumination estimation to the entire video is computationally expensive, we thus propose to obtain temporally coherent illumination sequence by propagating illuminations of sparsely sampled keyframes to the others. Fig.~\ref{fig:video_pipeline} shows the pipeline of our method for enhancing underexposed videos. For a given underexposed video, we first sample some keyframes, and then perform illumination estimation to obtain their illuminations. Next, we propagate these illuminations to other temporally adjacent frames. Finally, a video denoising operation is applied to remove noise in the enhanced video recovered from the obtained illumination sequence. In the following we describe each step in details.

\subsection{Keyframe Extraction \& Illumination Estimation}
The first step in our underexposed video enhancement pipeline searches for keyframes. Intuitively, keyframes that approximately depict the overall illumination changes of the source video are required to allow reliable illumination propagation. Based on this observation, we begin by taking the first frame as a keyframe, and then select the nearest frame that differs the first keyframe in luminance over $30\%$ pixels as the second keyframe. The third keyframe is similarly determined based on the second keyframe. We iteratively perform above operation to collect all keyframes. Note we compute the luminance difference in Lab color space, and consider two pixels to be different in luminance if the normalized difference is no less than a threshold $\ell=0.1$. In addition, a Gaussian smoothing is applied to the luminance channel of the source underexposed video to reduce the effect of noise before extracting the keyframes.

The second step in our pipeline estimates illumination for the collected keyframes. In order to achieve higher efficiency, the illumination estimation optimization in Eq.~\ref{equ:obj_func} together with the efficient implementation in Eq.~\ref{equ:jbu} are employed to obtain illumination for each keyframe.

\begin{figure}
  \centering
  \includegraphics[width=2.8in]{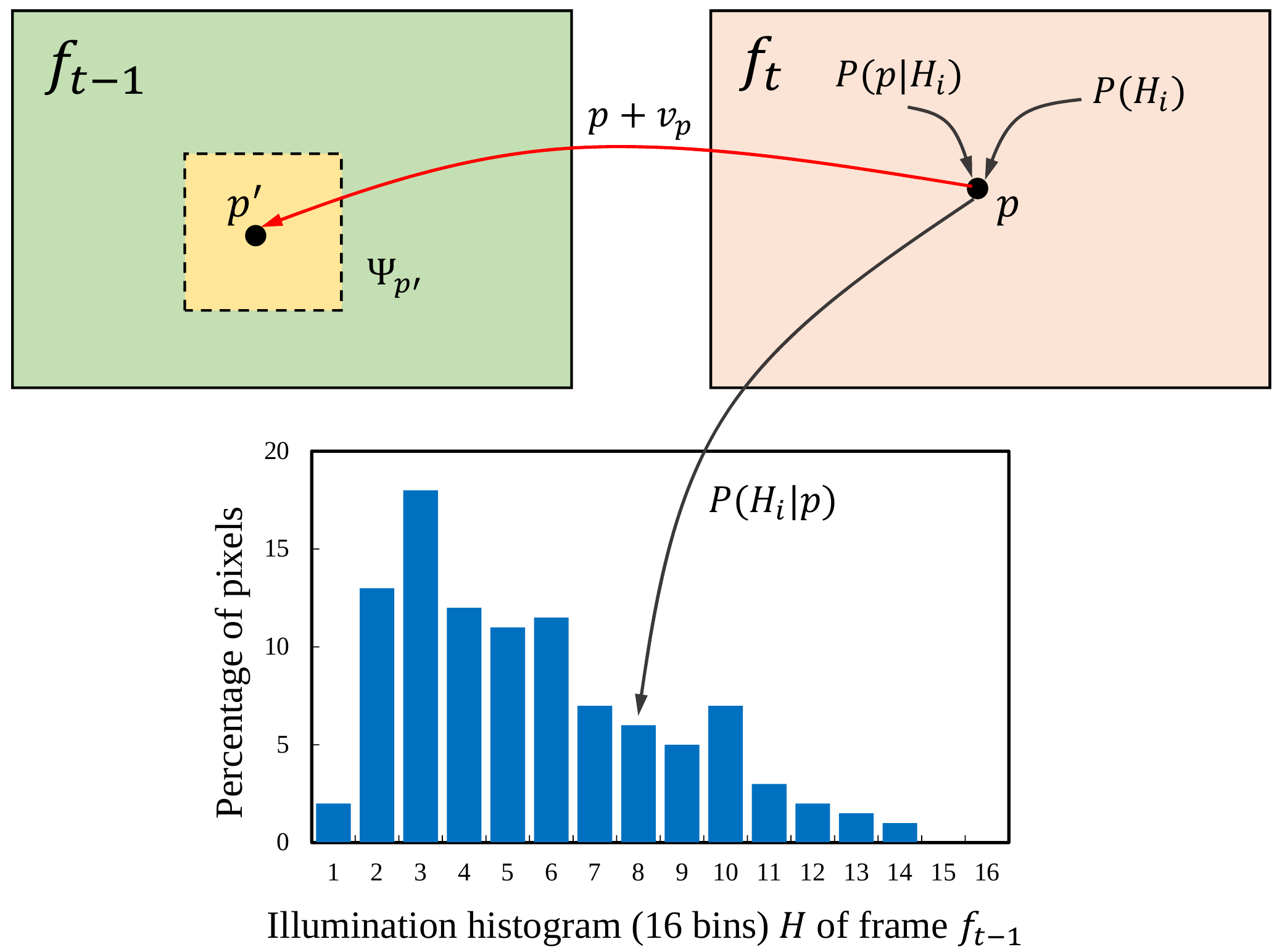}
  \caption{Illustration of the illumination propagation between two consecutive frames $f_{t-1}$ (with known illumination) and $f_t$ (illumination is unknown). For a pixel $p$ in frame $f_t$, we first find its corresponding pixel $p'=p+v_p$ in frame $f_{t-1}$ using the estimated motion flow $v_p$. Based on the illumination histogram $H$ of frame $f_{t-1}$, we then predict $P(p|H_i)$ by exploring the luminance similarity between pixel $p$ and the pixels within a squared window $\Psi_{p'}$ centered at $p'$, and approximate $P(H_i)$ based on the spatial proximity of pixels within the window $\Psi_{p'}$. With $P(p|H_i)$ and $P(H_i)$, we can finally find the histogram bin of $H$ that pixel $p$ most likely belongs to by tackling a MAP problem in Eq.~\ref{equ:map}.}\label{fig:propagation}
  \vspace{-2mm}
\end{figure}

\subsection{Temporal Illumination Propagation}
In the third step, we propagate illuminations of the keyframes to the rest of video frames. For each keyframe, we propagate its illumination over successive frames, until a new keyframe is found to start a new round of illumination propagation. We iteratively implement above illumination propagation until we reach the end of the video. Fig.~\ref{fig:propagation} shows how our illumination propagation works.

\begin{figure*}
	\centering
	\captionsetup[subfigure]{labelformat=empty}
	\begin{subfigure}[c]{0.23\textwidth}
		\centering
		\includegraphics[width=1.65in]{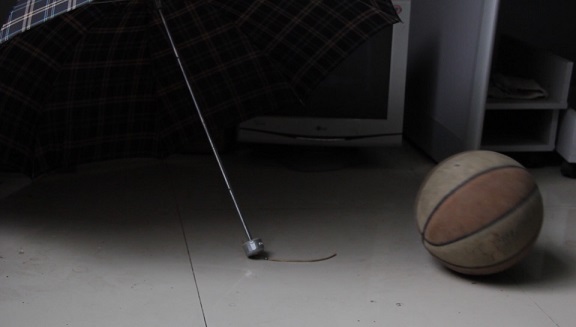} \\ \vspace{1mm}
		\includegraphics[width=1.65in]{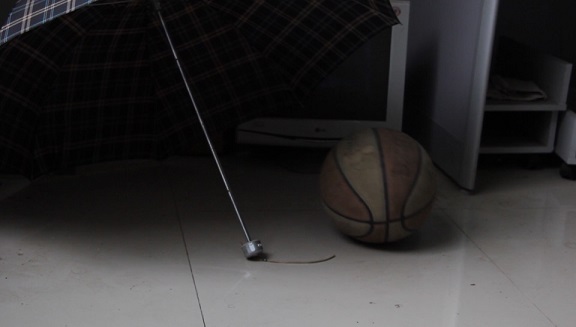}
		\caption{Source video}
	\end{subfigure}
	\begin{subfigure}[c]{0.23\textwidth}
		\centering
		\includegraphics[width=1.65in]{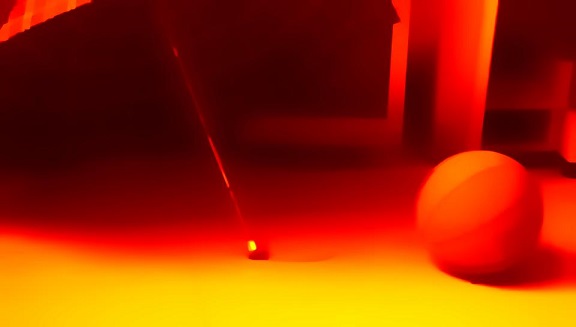} \\ \vspace{1mm}
		\includegraphics[width=1.65in]{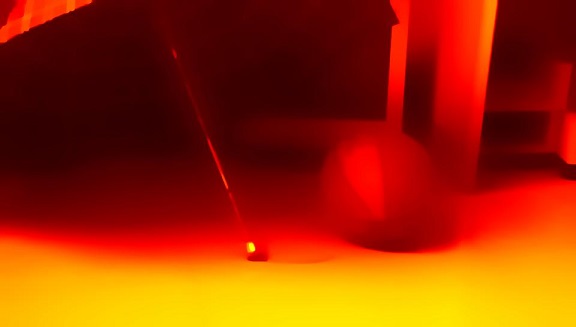}
		\caption{Illumination sequence}
	\end{subfigure}
	\begin{subfigure}[c]{0.23\textwidth}
		\centering
		\includegraphics[width=1.65in]{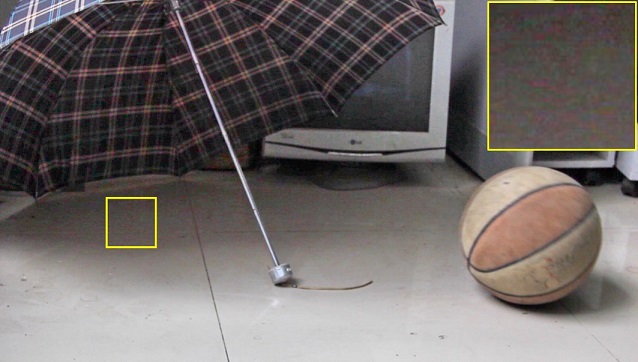} \\ \vspace{1mm}
		\includegraphics[width=1.65in]{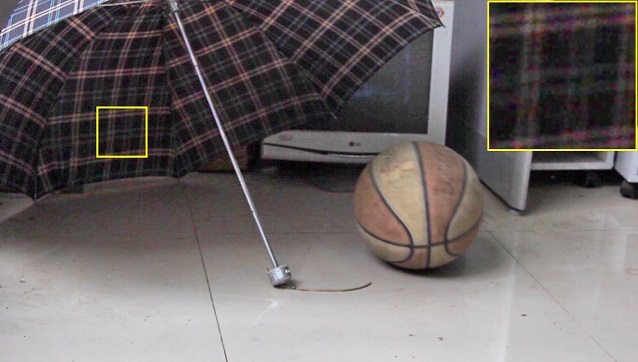}
		\caption{Enhanced video}
	\end{subfigure}
	\begin{subfigure}[c]{0.23\textwidth}
		\centering
		\includegraphics[width=1.65in]{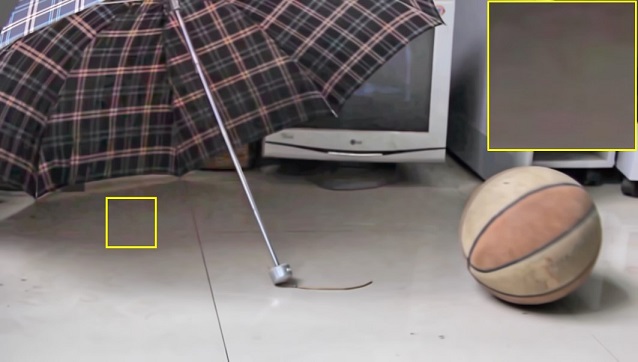} \\ \vspace{1mm}
		\includegraphics[width=1.65in]{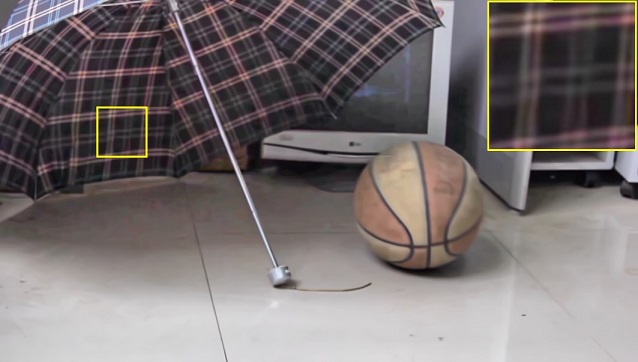}
		\caption{Final denoised video}
	\end{subfigure}
	\caption{An example underexposed video enhanced by our approach. The $1^{st}$ and $16^{th}$ frames are shown here.}
	\label{fig:video_example}
	\vspace{-2mm}
\end{figure*}

Let $f_t (t = 1, 2...)$ be the frames of an input video. For a pixel $p$ in frame $f_t$ (not a keyframe and the illumination is unknown at this point), with the luminance channel (i.e., Y channel in YUV color space) as $L_{p, t}$, we aim to predict its most likely illumination value based on the previous frame $f_{t-1}$ (either a keyframe or a frame with propagated known illumination) using a Bayesian formulation. To simplify the problem, we construct a histogram $H$ of 16 bins for illumination values of the frame $f_{t-1}$, where $H_i$ denotes the $i$-th bin and $|H_i|$ returns the number of pixels assigned to the bin. In this way, the illumination propagation problem reduces to finding the histogram bin of the previous frame $f_{t-1}$ that pixel $p$ in current frame $f_t$ belongs to. To achieve this, we introduce a probabilistic approach to find the bin index that maximizes the posterior probability $P(H_i|p)$ by addressing a Maximum A Posteriori (MAP) problem as
\begin{equation} \label{equ:map}
i = \mathop{\arg \max}_{i} P(H_i|p) \propto P(p|H_i) P(H_i),
\end{equation}
where $P(p|H_i)$ denotes the likelihood that pixel $p$ belongs to the bin $H_i$. $P(H_i)$ is a prior. Below we describe these two terms in detail.

We compute the likelihood of illumination value of a pixel $p$ in frame $f_t$ that belongs to the bin $H_i$ based on the probability density function of $H_i$. Our main idea is to employ standard non-parametric density estimation for calculating the compatibility of assigning a pixel to a bin. By adopting the Parzen window-based approximation, we define $P(p|H_i)$ as
\begin{equation}
P(p|H_i) = \frac{1}{|H_i|}\sum_{q \in \Psi_{p'}}\phi_i(L_{q,t-1})\mathcal{G}\left(\frac{L_{q,t-1} - L_{p,t}}{d}\right),
\end{equation}
where $\Psi_{p'}$ denotes a $N \times N$ ($N = 30$) squared window centered at pixel $p'$ in frame $f_{t-1}$. $p' = p + v_p$ is the corresponding pixel of $p$ (in frame $f_t$), which is indicated by the motion vector $v_p$ between $f_t$ and $f_{t-1}$. $q$ indexes pixels within the window $\Psi_{p'}$. $L_{q,t-1}$ denotes the luminance value of the pixel $q$ in frame $f_{t-1}$, and $\phi_i(L_{q,t-1})$ returns the number of pixels with luminance value $L_{q,t-1}$ in the $i$-th bin $H_i$. $\mathcal{G}$ is a Parzen window defined by a 1-D Gaussian kernel function with width $d = 5$. Note the optical flow of the source video is computed by the method of \cite{brox2004high}.

Explicitly computing $P(H_i)$ is difficult, we instead follow common MAP solutions~\cite{ye2014intrinsic} to devise a smoothness term to approximate the prior $P(H_i)$. For the pixel $p'$ ($f_{t-1}$) computed from $p$ ($f_t$) by optical flow, we define $D(p',q')$ as the Euclidean distance between pixels $p'$ and $q'$, where $q'$ denotes a pixel within a squared window centered at $p'$ that belongs to the $i$-th bin $H_i$. Formally, the prior $P(H_i)$ is formulated as
\begin{equation}
P(H_i) = \min_{q'}\left(\frac{1}{\sqrt{D(p', q')}}\right).
\end{equation}
Note $P(H_i)$ is feasible to be treated as a prior since it is irrelevant to the luminance of a pixel, and can be computed after the previous frame has been processed.

\subsection{Video Denoising}
While the proposed method can robustly enhance underexposed videos, it may also amplify the underlying noise. Unlike still images, the noise issue is usually non-negligible for dynamic video. Thus, to further improve the visual quality, we in the final step employ a video denoising operation to reduce the noise level of the enhanced video. In order to trade off the denoising performance and the runtime efficiency, we adopt V-BM4D \cite{maggioni2012video}, though any other video denoising algorithms would also work with our method.

\begin{figure}
	\centering
	\begin{subfigure}[c]{0.22\textwidth}
		\centering
		\includegraphics[width=1.57in]{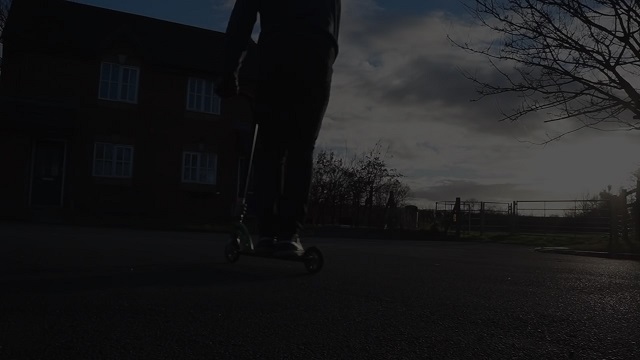} \\ \vspace{1mm}
		\includegraphics[width=1.57in]{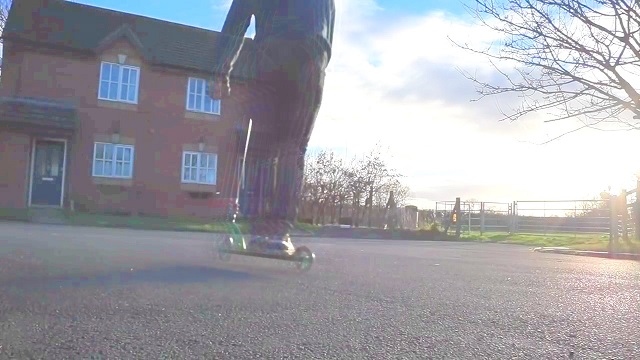} \\ \vspace{1mm}
		\includegraphics[width=1.57in]{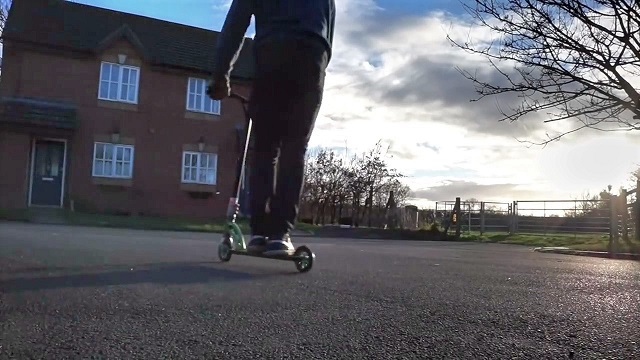} \\ \vspace{1mm}
		\includegraphics[width=1.57in]{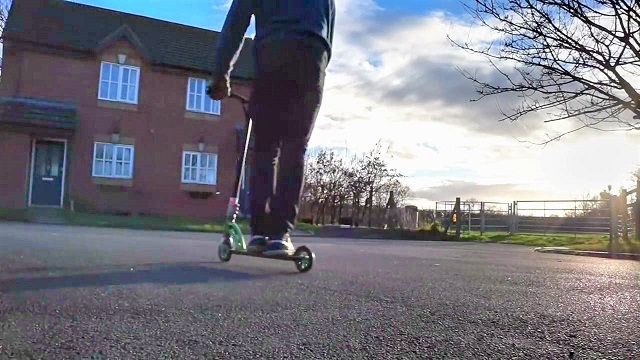}
	\end{subfigure}
	\begin{subfigure}[c]{0.22\textwidth}
		\centering
		\includegraphics[width=1.57in]{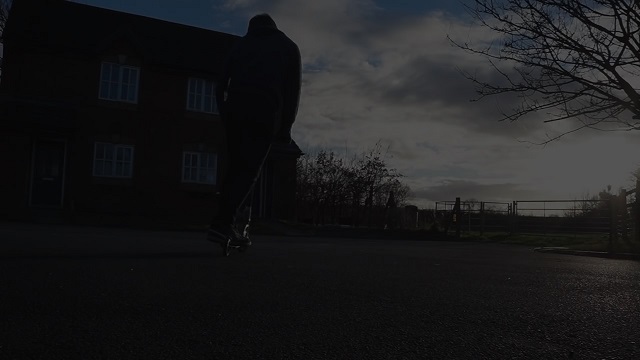} \\ \vspace{1mm}
		\includegraphics[width=1.57in]{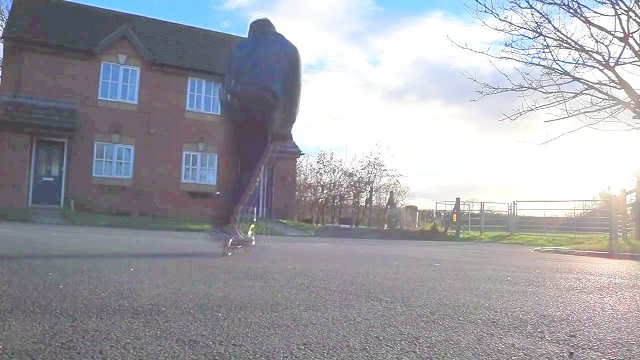} \\ \vspace{1mm}
		\includegraphics[width=1.57in]{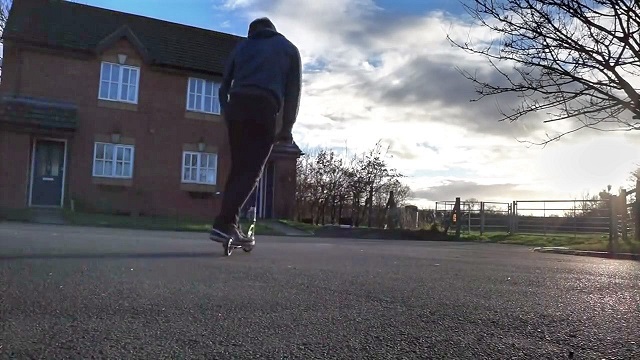} \\ \vspace{1mm}
		\includegraphics[width=1.57in]{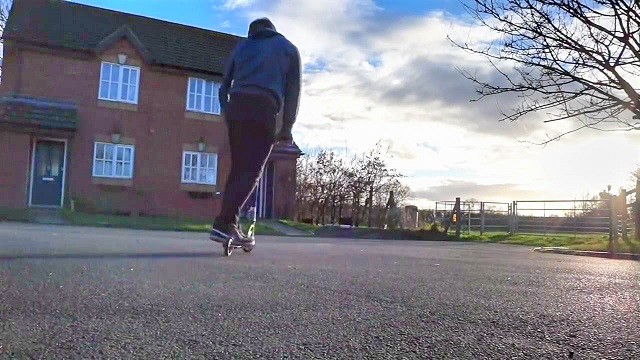}
	\end{subfigure}
	\caption{Comparison with exiting methods on enhancing a challenging underexposed video. From top to bottom are the input, result of~\cite{bennett2005video},~\cite{zhang2016underexposed} and our method. }
	\label{fig:video_compare}
    \vspace{-2mm}
\end{figure}

\begin{figure*}
	\centering
	\begin{subfigure}[c]{0.23\textwidth}
		\centering
		\includegraphics[width=1.65in]{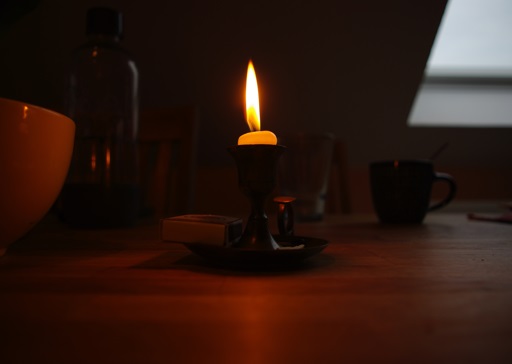}
		\caption{Input}
	\end{subfigure}
	\begin{subfigure}[c]{0.23\textwidth}
		\centering
		\includegraphics[width=1.65in]{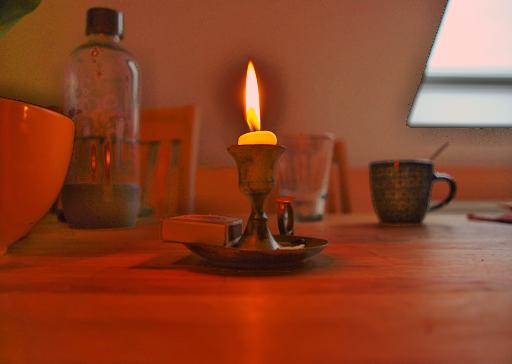}
		\caption{NPE~\cite{wang2013naturalness}}
	\end{subfigure}
	\begin{subfigure}[c]{0.23\textwidth}
		\centering
		\includegraphics[width=1.65in]{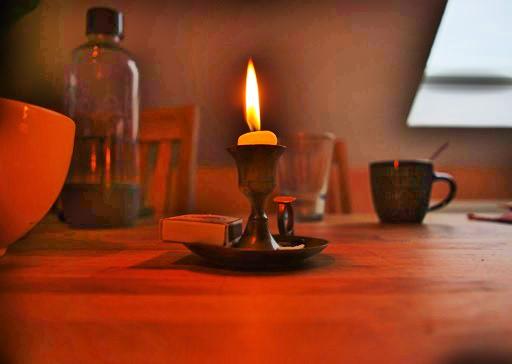}
		\caption{WVM~\cite{fu2016weighted}}
	\end{subfigure}
	\begin{subfigure}[c]{0.23\textwidth}
		\centering
		\includegraphics[width=1.65in]{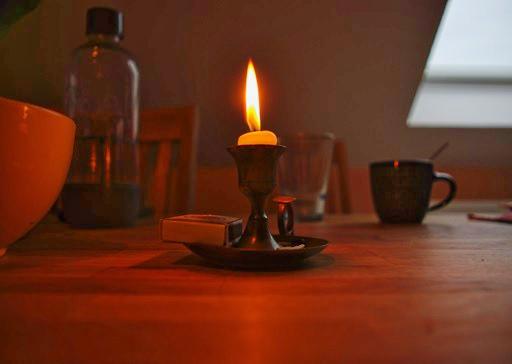}
		\caption{JieP~\cite{cai2017joint}}
	\end{subfigure}  \\ \vspace{2mm}
	\begin{subfigure}[c]{0.23\textwidth}
		\centering
		\includegraphics[width=1.65in]{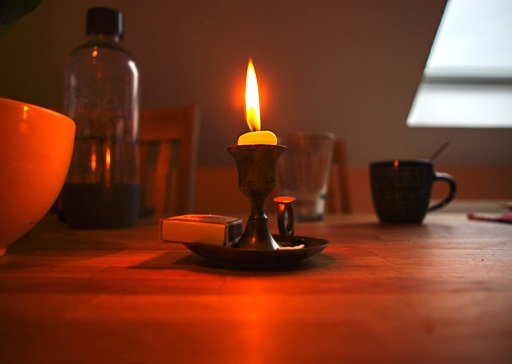}
		\caption{LIME~\cite{guo2017lime}}
	\end{subfigure}
	\begin{subfigure}[c]{0.23\textwidth}
		\centering
		\includegraphics[width=1.65in]{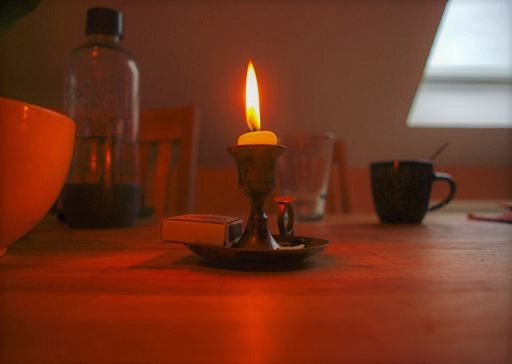}
		\caption{HDRNet~\cite{gharbi2017deep}}
	\end{subfigure}
	\begin{subfigure}[c]{0.23\textwidth}
		\centering
		\includegraphics[width=1.65in]{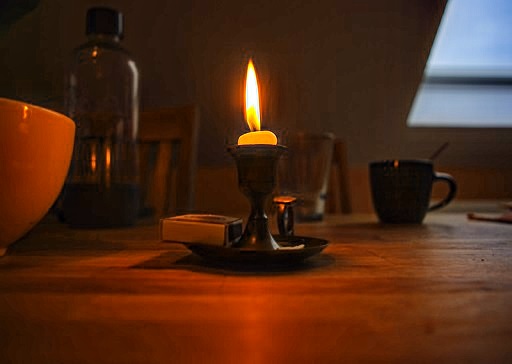}
		\caption{DPE~\cite{chen2018deep}}
	\end{subfigure}
	\begin{subfigure}[c]{0.23\textwidth}
		\centering
		\includegraphics[width=1.65in]{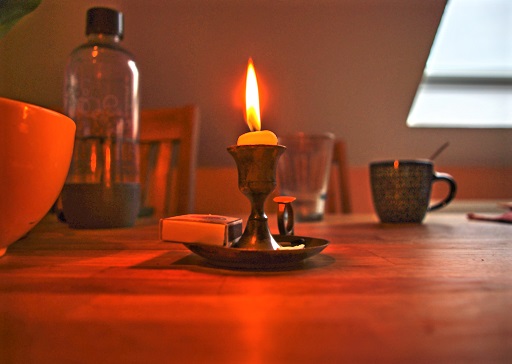}
		\caption{Ours}
	\end{subfigure}
	\caption{Visual comparison with state-of-the-art methods on a test image from the MEF dataset~\cite{ma2015perceptual}. }
	\label{fig:visual_compare_1}
	\vspace{-2mm}
\end{figure*}

\begin{figure*}
	\centering
	\begin{subfigure}[c]{0.23\textwidth}
		\centering
		\includegraphics[width=1.65in]{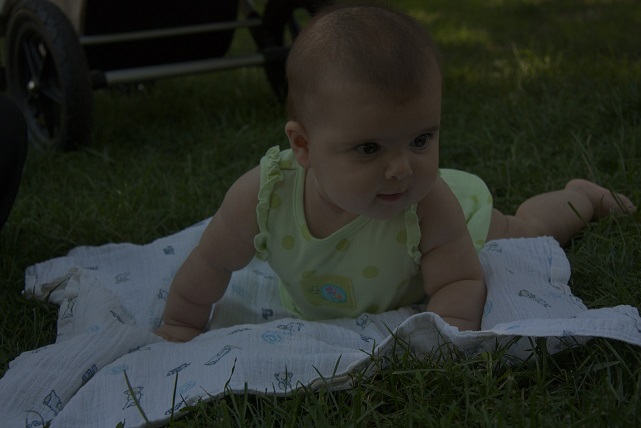}
		\caption{Input}
	\end{subfigure}
	\begin{subfigure}[c]{0.23\textwidth}
		\centering
		\includegraphics[width=1.65in]{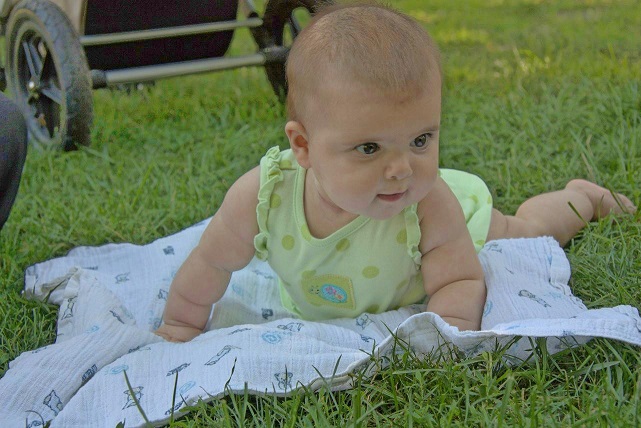}
		\caption{NPE~\cite{wang2013naturalness}}
	\end{subfigure}
	\begin{subfigure}[c]{0.23\textwidth}
		\centering
		\includegraphics[width=1.65in]{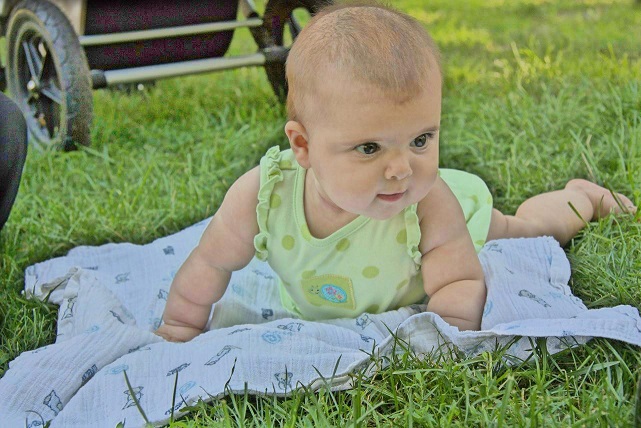}
		\caption{WVM~\cite{fu2016weighted}}
	\end{subfigure}
	\begin{subfigure}[c]{0.23\textwidth}
		\centering
		\includegraphics[width=1.65in]{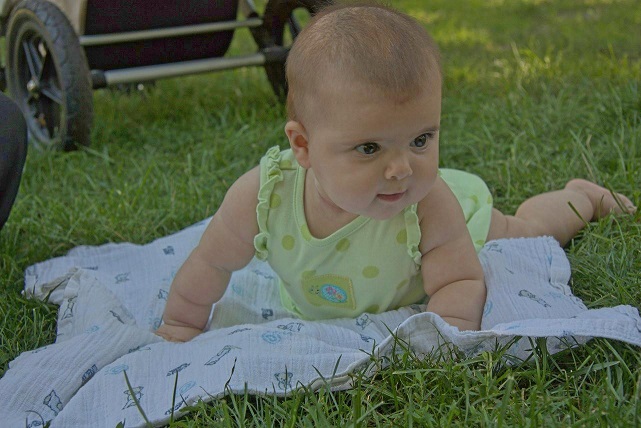}
		\caption{JieP~\cite{cai2017joint}}
	\end{subfigure}  \\ \vspace{2mm}
	\begin{subfigure}[c]{0.23\textwidth}
		\centering
		\includegraphics[width=1.65in]{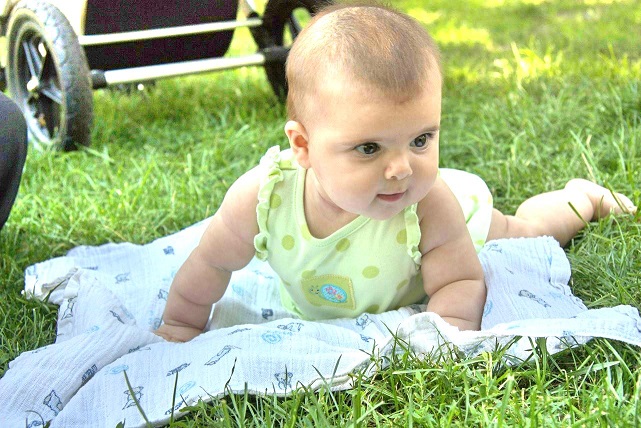}
		\caption{LIME~\cite{guo2017lime}}
	\end{subfigure}
	\begin{subfigure}[c]{0.23\textwidth}
		\centering
		\includegraphics[width=1.65in]{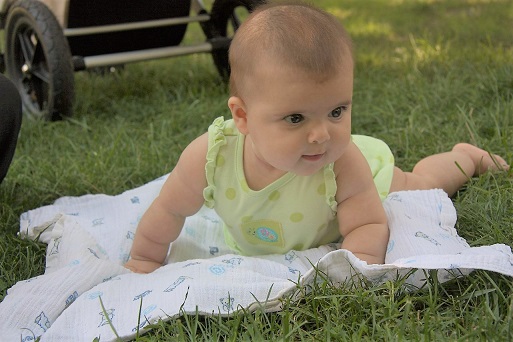}
		\caption{HDRNet~\cite{gharbi2017deep}}
	\end{subfigure}
	\begin{subfigure}[c]{0.23\textwidth}
		\centering
		\includegraphics[width=1.65in]{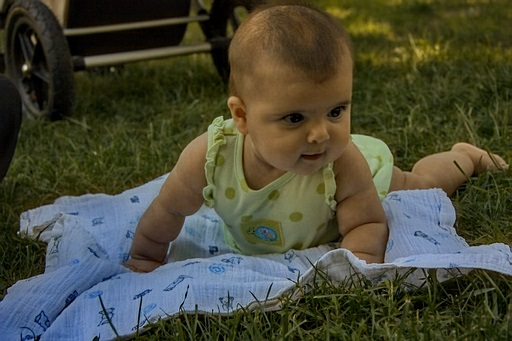}
		\caption{DPE~\cite{chen2018deep}}
	\end{subfigure}
	\begin{subfigure}[c]{0.23\textwidth}
		\centering
		\includegraphics[width=1.65in]{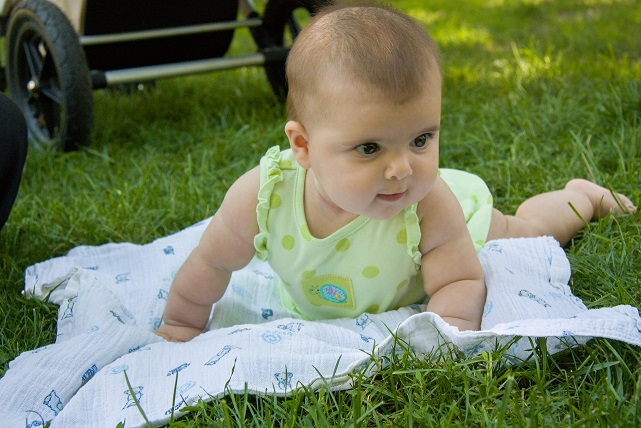}
		\caption{Ours}
	\end{subfigure}
	\caption{Visual comparison with state-of-the-art methods on a test image from the FiveK dataset~\cite{bychkovsky2011learning}.}
	\label{fig:visual_compare_2}
	\vspace{-2mm}
\end{figure*}

\begin{figure}
    \centering
    \includegraphics[width=2.1in]{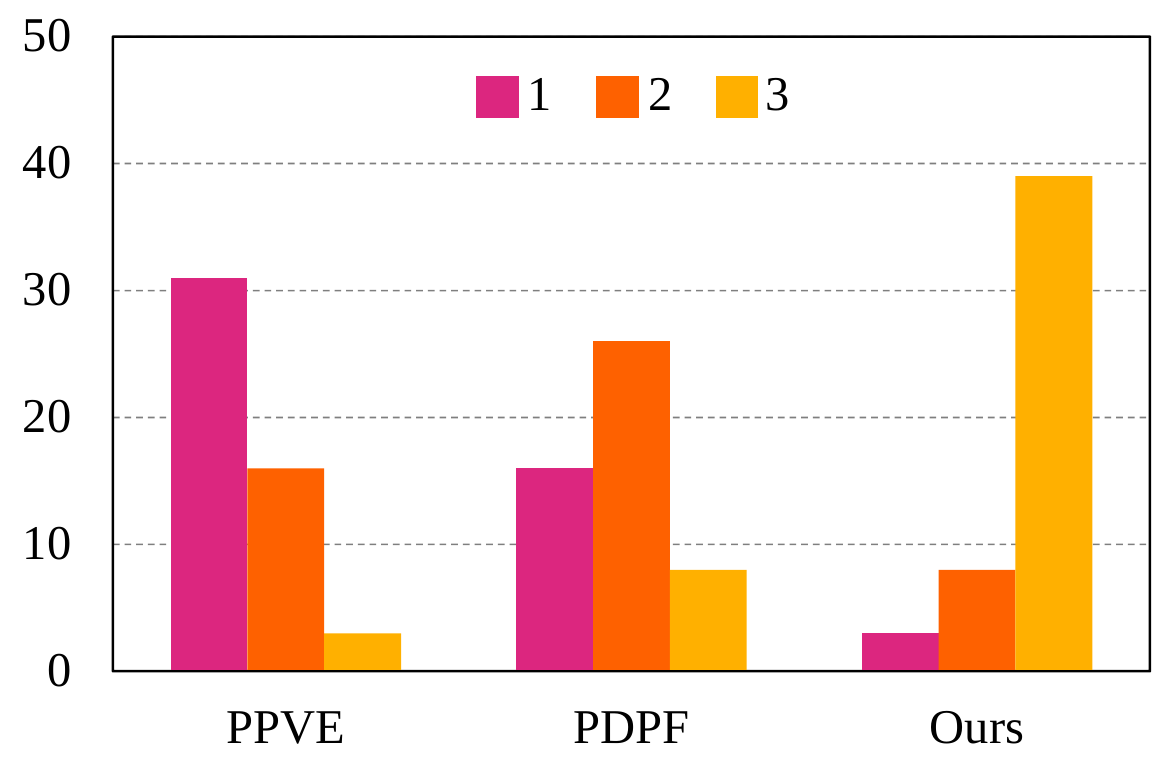}
    \caption{Rating distributions for video enhancement methods in the user study. Higher ratings indicate better results.}
    \label{fig:video_user_study}
    \vspace{-4mm}
\end{figure}

\subsection{Result and Comparison}
Fig.~\ref{fig:video_example} shows an example underexposed video enhanced by our approach. As can be seen, by obtaining the illumination sequence, we successfully light up the underexposed regions and reveal the underlying texture details of the umbrella. The video denoising operation further reduces the noise level of the enhanced video and generates a better result. Fig.~\ref{fig:video_compare} compares our method with previous underexposed video enhancement methods PPVE \cite{bennett2005video} and PDPF \cite{zhang2016underexposed}. We can see that \cite{bennett2005video} produces over-saturated result and induces clear jittering artifacts around the legs, while result of \cite{zhang2016underexposed} fails to present distinct contrast and vivid color. In comparison, our method produces a more appealing result. Note that the average counts of frames between two adjacently sampled keyframes for the videos shown in Fig.~\ref{fig:video_example} and Fig.~\ref{fig:video_compare} are 25 and 19.

\begin{table}
\centering
\vspace{-2mm}
\caption{Quantitative comparison between our method and other video enhancement methods in terms of ``mean/standard deviation'' of DE and NIQE.}
\begin{tabular}{l|c|c|c|c}
\hline
     & Input     & PPVE~\cite{bennett2005video}      & PDPF~\cite{zhang2016underexposed}      & Ours      \\ \hline
DE   & 5.81/0.27 & 7.38/0.38 & 7.14/0.31 & 7.53/0.23 \\ \hline
NIQE & 4.35/0.41 & 3.63/0.47 & 3.37/0.43 & 3.12/0.35 \\ \hline
\end{tabular} \label{table:quantitative_video}
\vspace{-4mm}
\end{table}

\begin{table*}[]
	\centering
	\caption{Quantitative comparison between our method and the state-of-the-arts on the six employed datasets.}
	\label{my-label}
	\begin{tabular}{l|cc|cc|cc|cc|cc|cc|cc|cc}
		\hline
		\multirow{2}{*}{Dataset} & \multicolumn{2}{c|}{Original} & \multicolumn{2}{c|}{NPE~\cite{wang2013naturalness}} & \multicolumn{2}{c|}{WVM~\cite{fu2016weighted}} & \multicolumn{2}{c|}{JieP~\cite{cai2017joint}} & \multicolumn{2}{c|}{LIME~\cite{guo2017lime}} & \multicolumn{2}{c|}{HDRNet~\cite{gharbi2017deep}} & \multicolumn{2}{c|}{DPE~\cite{chen2018deep}} & \multicolumn{2}{c}{Ours}        \\
		& DE            & NIQE          & DE          & NIQE        & DE          & NIQE       & DE     & NIQE            & De         & NIQE       & DE          & NIQE        & DE           & NIQE        & DE             & NIQE           \\ \hline
		NPE                 & 6.56         & 3.89         & 7.22        & 3.18        & 7.03        & 3.55       & 7.34  & 3.11           & 7.54      & 3.31      & 7.33       & 3.51       & 7.13         & 3.62        & \textbf{7.64} & \textbf{3.02} \\
		MEF                 & 6.07         & 4.27         & 7.14        & 3.59        & 6.89        & 3.84       & 7.29  & 3.51           & 7.32     & 3.71      & 7.16       & 3.63       & 7.08         & 3.76        & \textbf{7.56} & \textbf{3.37} \\
		MF                  & 6.36         & 3.35         & 7.11        & 3.02        & 7.14        & 3.25       & 7.23  & 3.17           & 7.49      & 3.12      & 7.19       & 3.26       & 7.03         & 3.41        & \textbf{7.74} & \textbf{2.81} \\
		LIME                & 6.02         & 4.47         & 6.91        & 4.09       & 6.82        & 4.29       & 6.98  & 3.87           & 7.39      & 4.10      & 7.18       & 3.95       & 6.87         & 4.31        & \textbf{7.45} & \textbf{3.57} \\
		VV                  & 6.63         & 3.38         & 7.43        & \textbf{2.73}        & 7.32        & 2.97       & 7.48  & 2.81  & 7.53      & 2.89      & 7.62       & 2.92       & 7.46         & 3.17        & \textbf{7.81} & 2.75          \\
		FiveK & 6.45 & 3.29 & 7.09 & 2.93 & 7.03 & 3.12 & 7.16 & 2.82 & 7.21 & 2.88 & 7.11 & 2.79 & 6.93 & 3.17 & \textbf{7.25} & \textbf{2.68} \\
		\hline
	\end{tabular}
	\label{table:de_niqe}
\end{table*}

\begin{figure*}
	\centering
	\includegraphics[width=2.06in]{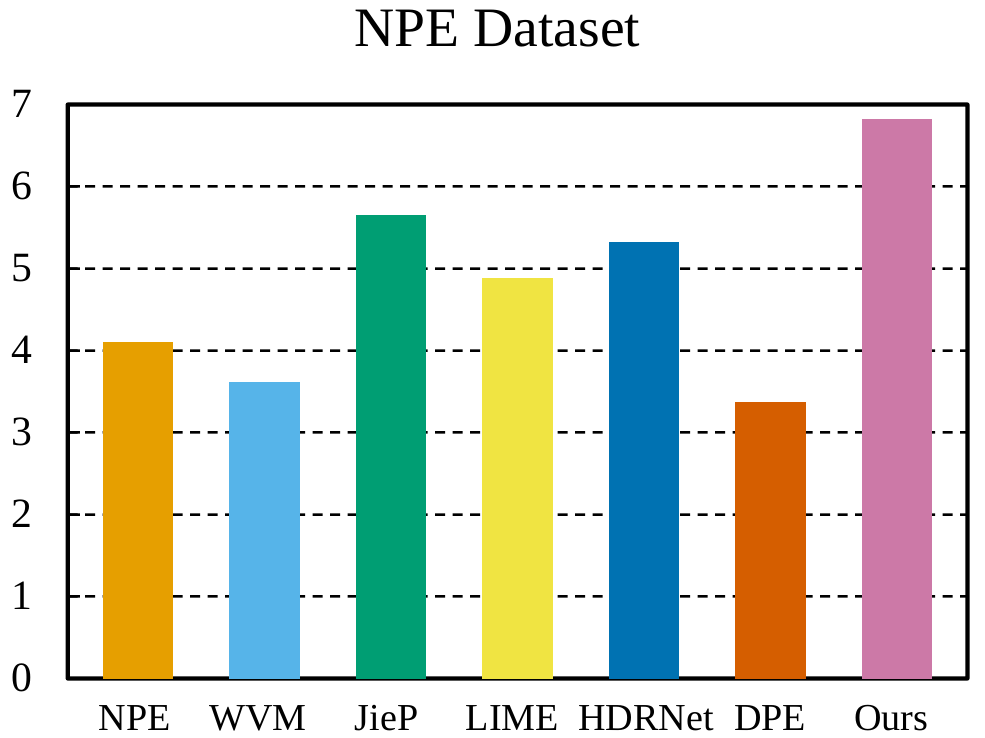} \hspace{2.6mm}
	\includegraphics[width=2.06in]{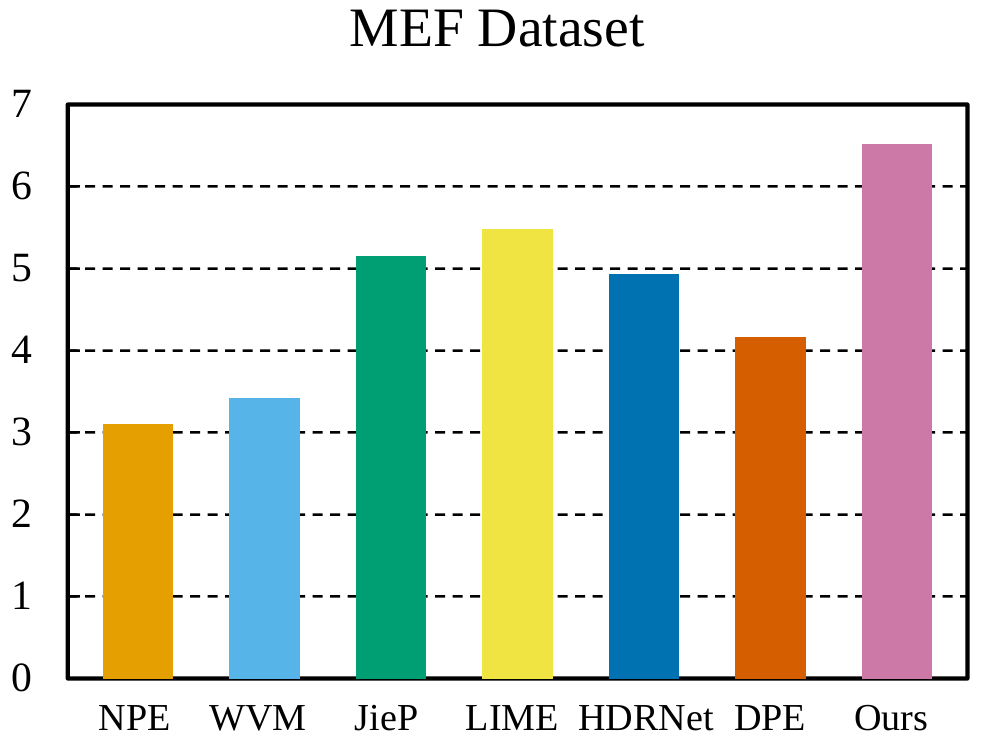} \hspace{2.6mm}
	\includegraphics[width=2.06in]{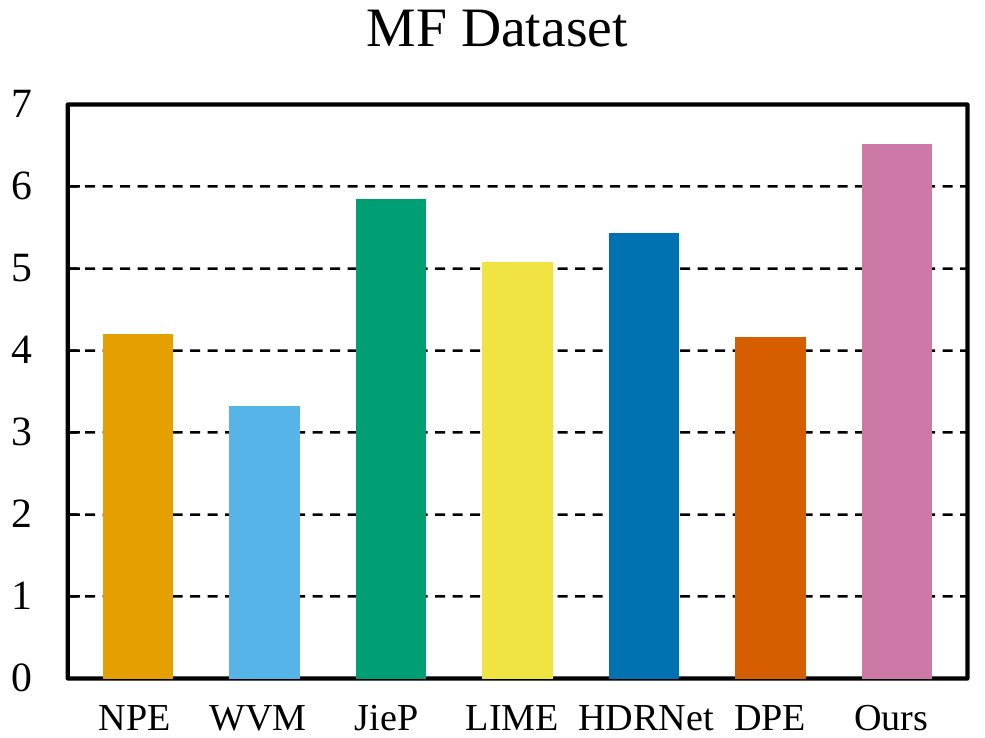} \\ \vspace{4mm}
	\includegraphics[width=2.06in]{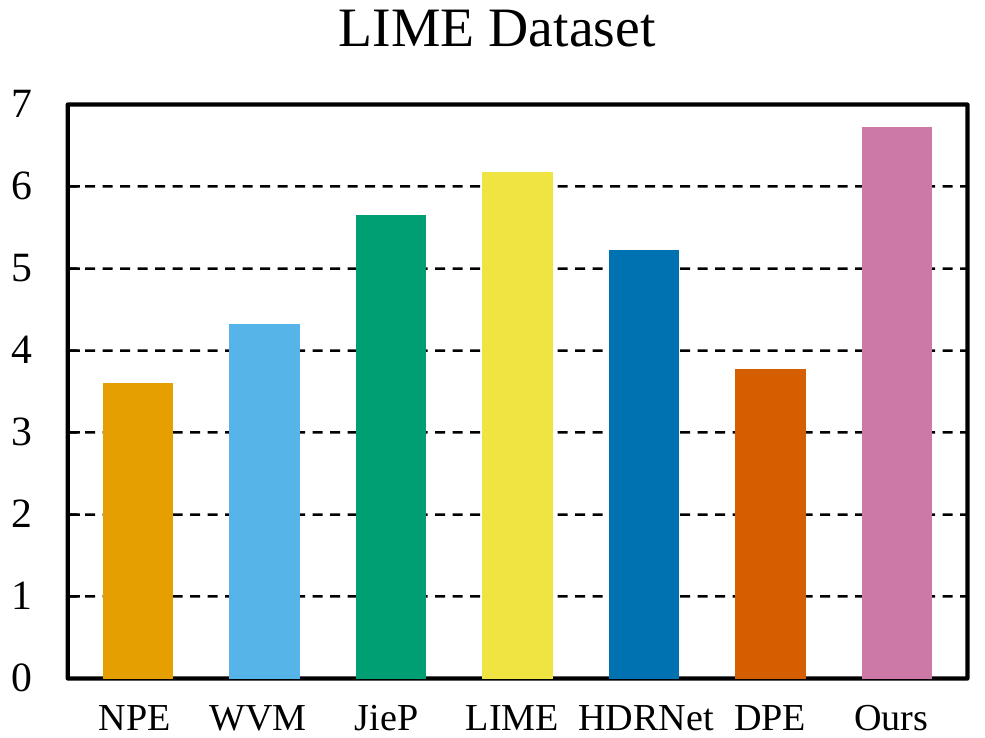} \hspace{2.6mm}
	\includegraphics[width=2.06in]{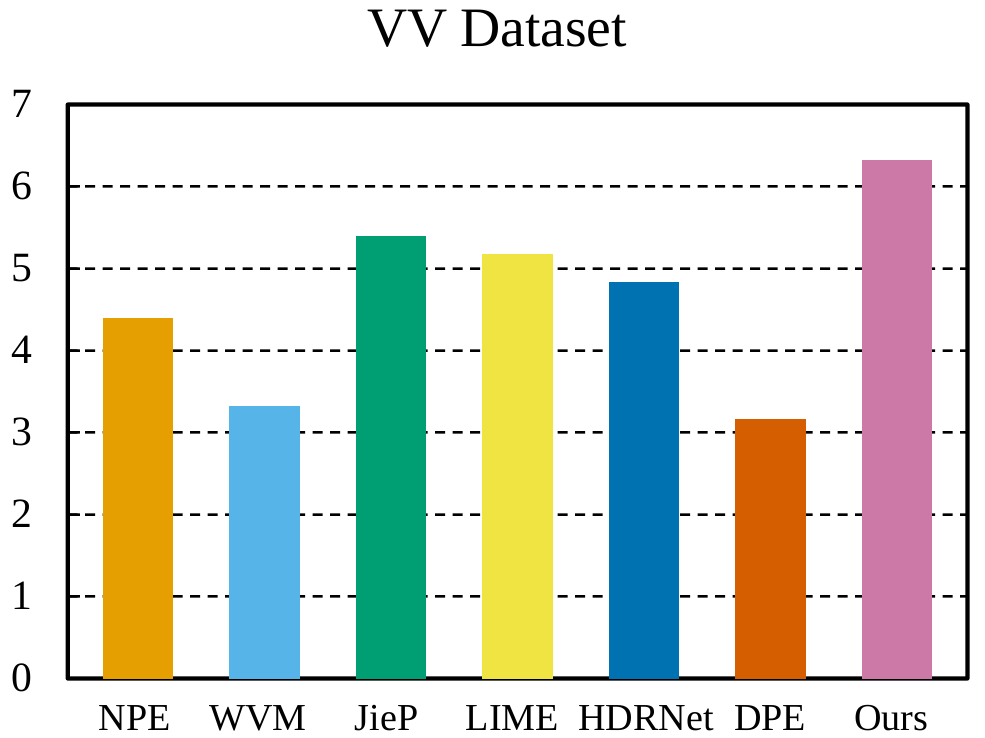} \hspace{2.6mm}
	\includegraphics[width=2.06in]{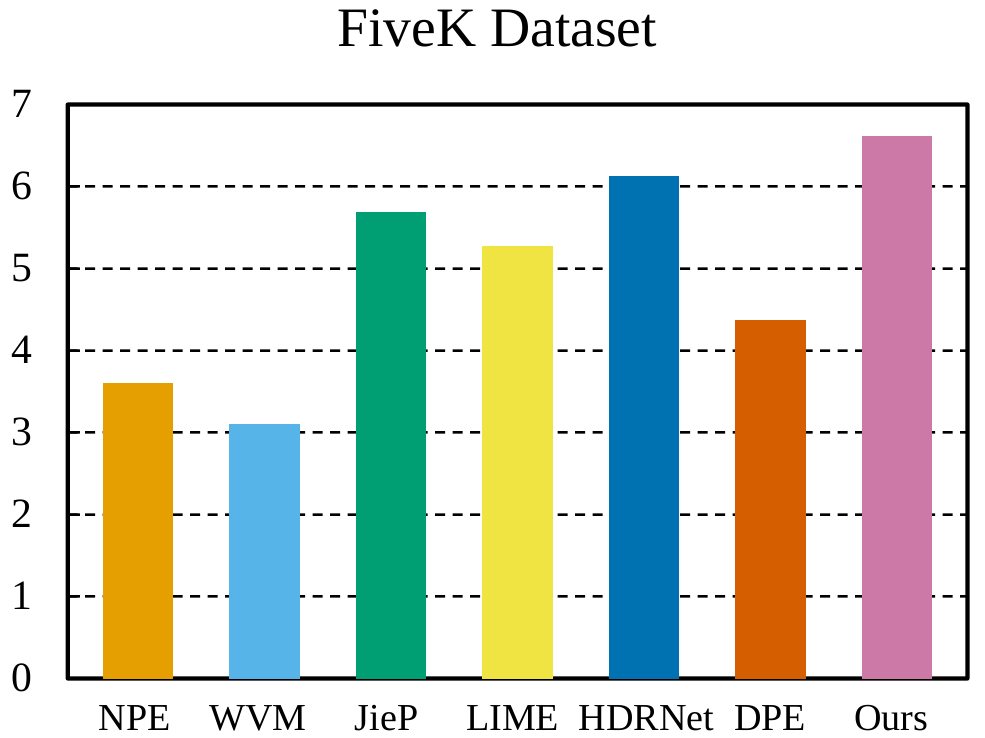}
	\caption{Ratings of different methods on the six employed datasets in the user study. The ordinate axis shows the average ratings received by the methods from the subjects on each dataset. Higher ratings indicate better results.}
	\vspace{-2mm}
	\label{fig:user_study}
\end{figure*}

We also follow~\cite{zhang2016underexposed} to evaluate video enhancement performance via user study. Specifically, we use five videos from~\cite{zhang2016underexposed} for testing. For each video, we ask 10 subjects to rank the enhancement results produced by~\cite{bennett2005video,zhang2016underexposed} and our method in terms of temporal consistency and visual effect using a rating scale from 1 (worst) to 3 (best). As shown in Fig.~\ref{fig:video_user_study}, the rating distribution shows that results produced by our method are more preferred by human subjects. Table~\ref{table:quantitative_video} further reports the DE and NIQE scores (see Section V.A for details of the two metrics) with mean and standard deviation for video enhancement results employed in the user study. As shown, our method outperforms the other two compared methods, since it achieves higher DE and lower NIQE values. Besides, our method also achieves lower standard deviation on the two metrics, demonstrating that it can better preserve the overall temporal consistency.

\section{Experiment}  \label{sec:experiment}

\subsection{Datasets and Evaluation Metrics}
\noindent \textbf{Benchmark datasets.} We employ six benchmark datasets to evaluate our method, which are the NPE dataset~\cite{wang2013naturalness}, MEF dataset~\cite{ma2015perceptual}, MF dataset~\cite{fu2016fusion}, LIME dataset~\cite{guo2017lime}, VV dataset \footnote{https://sites.google.com/site/vonikakis/datasets} and the FiveK dataset~\cite{bychkovsky2011learning}. Note that, for the FiveK dataset, we randomly select 100 underexposed images for evaluation, while the remaining 4900 images are used for training the HDRNet method \cite{gharbi2017deep} to be compared.

\vspace{0.3em}
\noindent \textbf{Evaluation metrics.} Since most benchmark datasets do not provide ground truth enhanced images, we employ two commonly-used non-reference metrics to quantitatively evaluate the algorithm performance. The first one is DE (discrete entropy)~\cite{ye2007discrete}, which measures the performance of detail/contrast enhancement. The second one is NIQE (natural image quality evaluator)~\cite{mittal2013making}, which is a learned model for assessing the overall naturalness of images. In general, high DE values of the enhanced images mean that the detail visibility of the original images are better improved, while low NIQE values indicate that the enhanced images own good naturalness. Although it is not absolutely true, high DE and low NIQE values usually indicate reasonably good results.

\subsection{Comparison with State-of-the-art Methods}
We compare our method with six recent underexposed photo enhancement methods: NPE~\cite{wang2013naturalness}, WVM~\cite{fu2016weighted}, JieP~\cite{cai2017joint}, LIME~\cite{guo2017lime}, HDRNet~\cite{gharbi2017deep} and DPE~\cite{chen2018deep}. The first four are Retinex-based methods, while the last two are deep-learning-based methods. For fair comparison, we obtain the results of the compared methods either from the online demo programs or by producing them using implementations provided by the authors with the recommend parameter setting. In the following, we conduct the comparison in three aspects, including visual comparison, quantitative comparison, and a user study.

\vspace{0.3em}
\noindent \textbf{Visual comparison.} We first show visual comparison in Fig.~\ref{fig:visual_compare_1} and~\ref{fig:visual_compare_2} on two challenging cases from the employed datasets: (i) a non-uniformly exposed photo with dim candlelight and imperceptible scene details (from the MEF dataset), (ii) an uniformly underexposed photo with little portrait details of the crawling baby (from the FiveK dataset). Comparing the results, we can see that our method outperforms the compared methods and has the following two advantages. First, it is able to recover more details and better contrast for the underexposed regions, without degrading other parts of the image. Second, it can reveal more vivid and natural colors, which makes our enhanced images look more realistic. Please see the supplementary material for more visual comparisons between our method and the state-of-the-arts.

\vspace{0.3em}
\noindent \textbf{Quantitative comparison.} Second, we quantitatively evaluate the performance of our method by comparing it with other methods in terms of the DE and NIQE metrics. Table~\ref{table:de_niqe} reports the quantitative comparison results. Note that, the original average DE and NIQE values for each dataset are also shown for reference. As can be seen, all methods increase the DE value due to the detail/contrast enhancement, and reduce the NIQE value because of lightening the underexposed regions. In contrast, our method achieves higher DE and lower NIQE than other compared methods on almost all the datasets, which manifests that our method can not only recover clearer details and more distinct contrast, but also better preserve the overall naturalness and photorealism of the enhanced images.

\begin{figure}
	\centering
	\begin{subfigure}[c]{0.148\textwidth}
		\centering
		\includegraphics[width=1.06in]{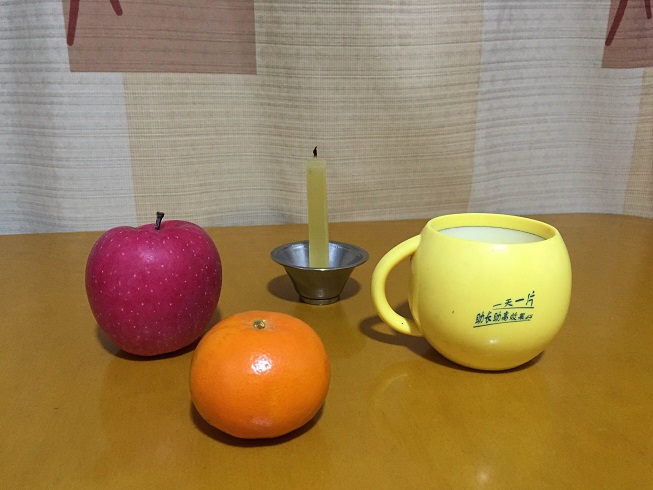}
		\caption{}
	\end{subfigure}
	\begin{subfigure}[c]{0.148\textwidth}
		\centering
		\includegraphics[width=1.06in]{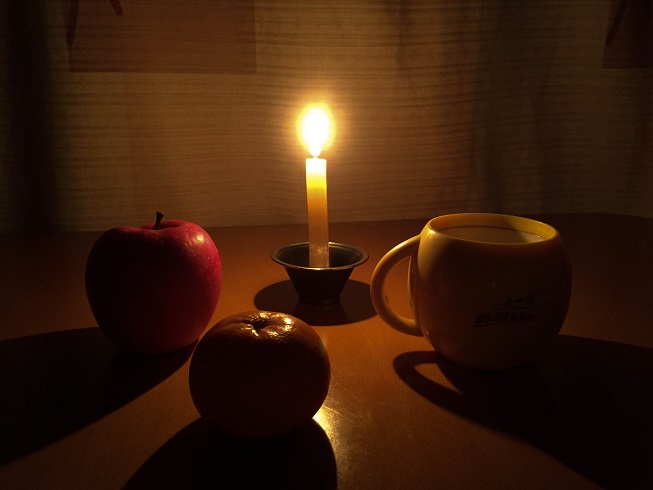}
		\caption{}
	\end{subfigure}
	\begin{subfigure}[c]{0.148\textwidth}
		\centering
		\includegraphics[width=1.06in]{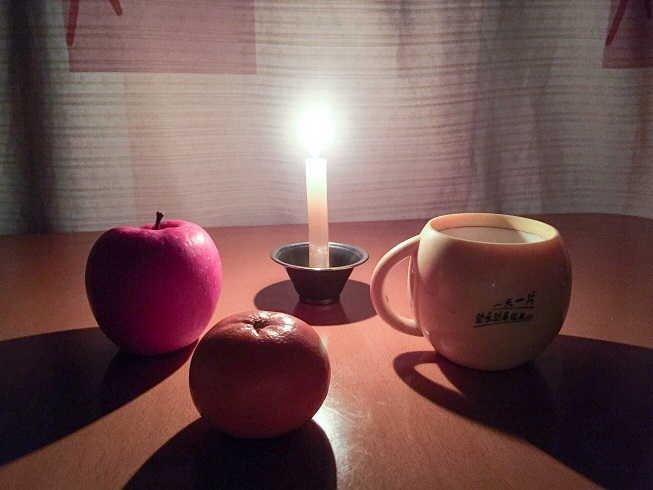}
		\caption{}
	\end{subfigure}	
	\caption{Effect of three-channel RGB illumination map in producing enhanced image with visually plausible color constancy effect. (a) A properly exposed image with a white artificial light. (b) A reference underexposed image captured by turning off the artificial light and lighting the candle instead. (c) Our enhance image for the underexposed image (b). }
	\label{fig:constancy}
	\vspace{-2mm}
\end{figure}

\begin{figure}
	\centering
	\begin{subfigure}[c]{0.225\textwidth}
		\centering
		\includegraphics[width=1.6in]{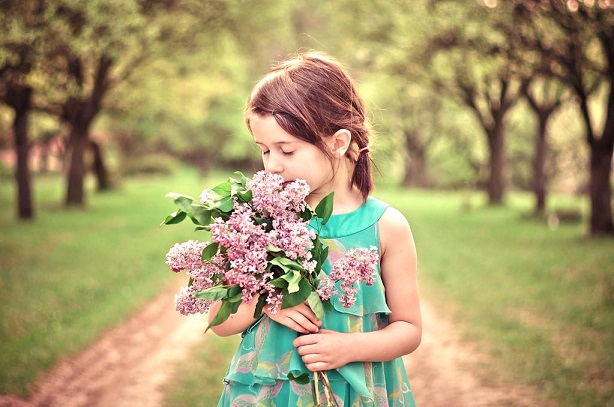} \\ \vspace{1mm}
		\includegraphics[width=1.6in]{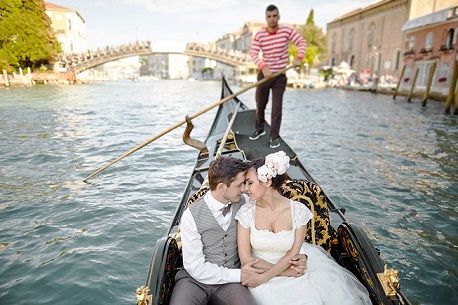}
		\caption{Input}
	\end{subfigure}
	\begin{subfigure}[c]{0.225\textwidth}
	\centering
		\includegraphics[width=1.6in]{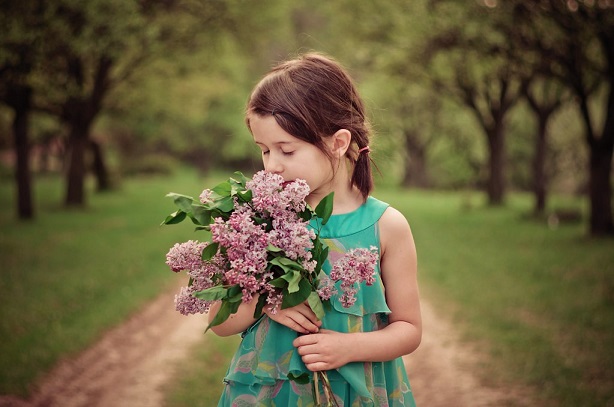} \\ \vspace{1mm}
		\includegraphics[width=1.6in]{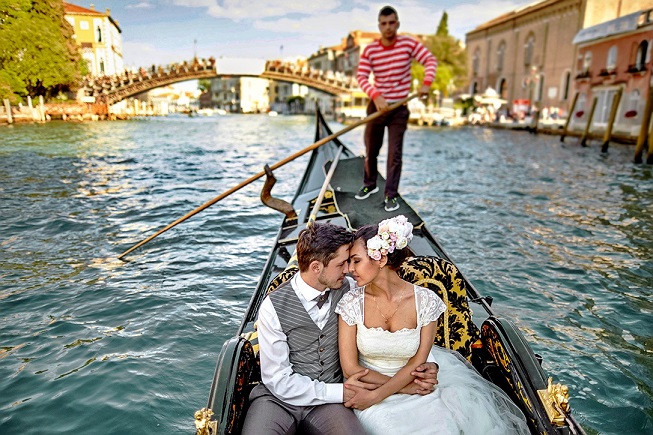}
		\caption{Our result}
	\end{subfigure}
	\caption{Two overexposed image corrected by our method.}
	\label{fig:overexposure}
    \vspace{-2mm}
\end{figure}

\vspace{0.3em}
\noindent \textbf{User study.} Since evaluating the visual quality of the enhanced images involves judgement of personal preference, we further conducted a user study to compare the results. To this end, we enhanced each test image in the six employed datasets using our method and the other six compared methods, and recruited 100 subjects via Amazon Mechanical Turk to rate the results. Specifically, for each test image, each subject was asked to rate seven different enhancement results (ours and other six methods') using a Likert scale from 1 (worst) to 7 (best), according to the following common requirements for the results: (i) clear details and distinct contrast, (ii) natural and vivid color, (iii) no loss of detail and overexposure, (iv) well-preserved photorealism. To avoid subjective bias, the subjects were assigned with anonymous results in random orders. After the subjects finished rating all the results, we computed the average ratings obtained by each method on different datasets. Fig.~\ref{fig:user_study} summarizes the ratings, where we can see that our method receives higher ratings compared to the others, demonstrating that results generated by our algorithm are more preferred by human subjects in average.

\subsection{More Analysis}
\noindent \textbf{Relationship to color constancy.} Our approach can also be extended to producing visually plausible color constancy effect by performing the illumination estimation separately on each RGB channel. As shown in Fig.~\ref{fig:constancy}, compared with the properly exposed image, our method not only improves the scene visibility of the underexposed image, but also partially removes the color of candlelight, e.g., the background curtain. Note that color constancy is a challenging problem, and low light condition would make the problem more difficult. Our three-channel illumination map extension is just a very simple trial to this problem. Hence, it may not always produce satisfactory color constancy effect, e.g., the desktop in Fig.~\ref{fig:constancy}(c).

\begin{figure}
	\centering
	\begin{subfigure}[c]{0.225\textwidth}
		\centering
		\includegraphics[width=1.61in]{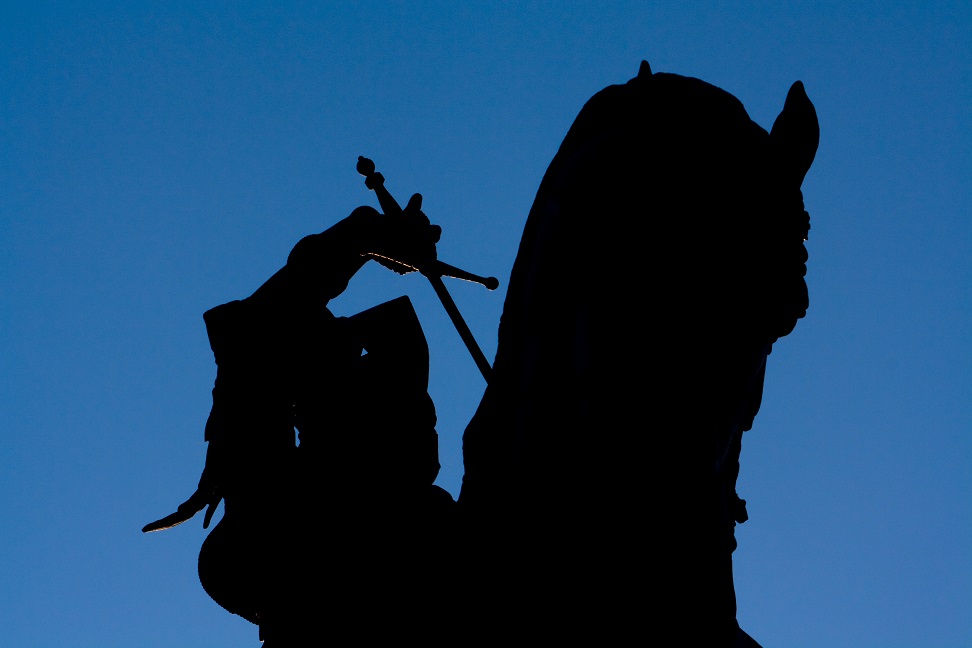}
		\caption{Input}
	\end{subfigure}
	\begin{subfigure}[c]{0.225\textwidth}
		\centering
		\includegraphics[width=1.61in]{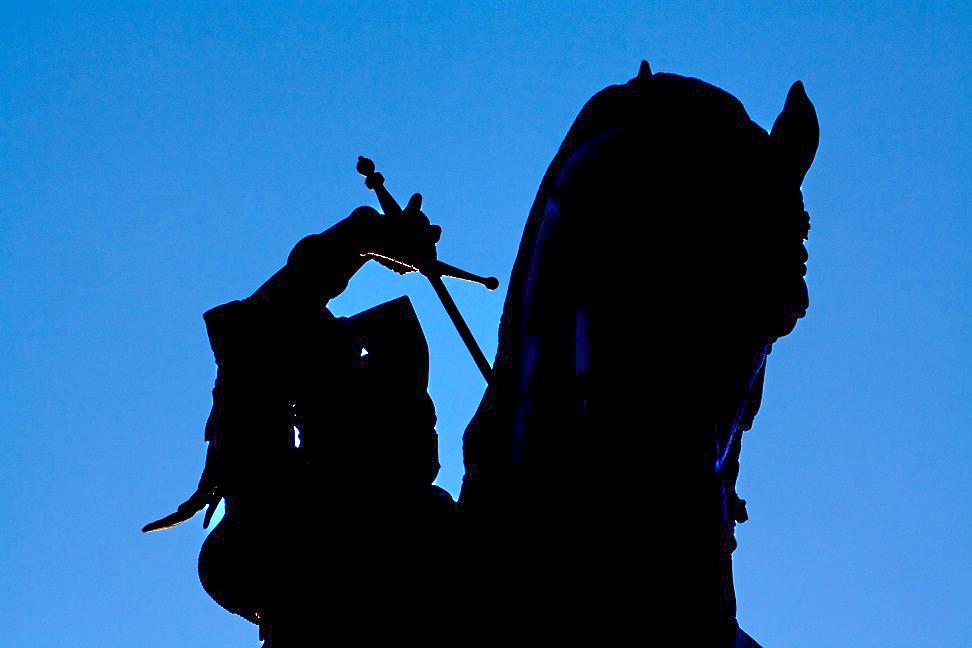}
		\caption{LIME~\cite{guo2017lime}}
	\end{subfigure} \\ \vspace{2mm}
	\begin{subfigure}[c]{0.225\textwidth}
		\centering
		\includegraphics[width=1.61in]{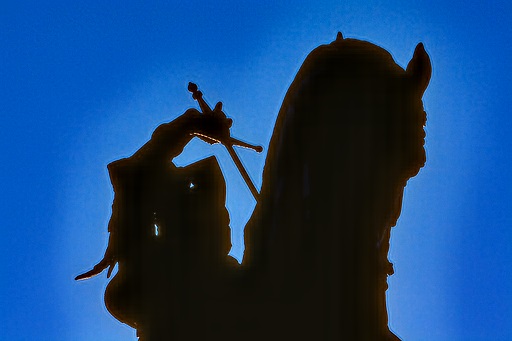}
		\caption{DPE~\cite{chen2018deep}}
	\end{subfigure}
	\begin{subfigure}[c]{0.225\textwidth}
		\centering
		\includegraphics[width=1.61in]{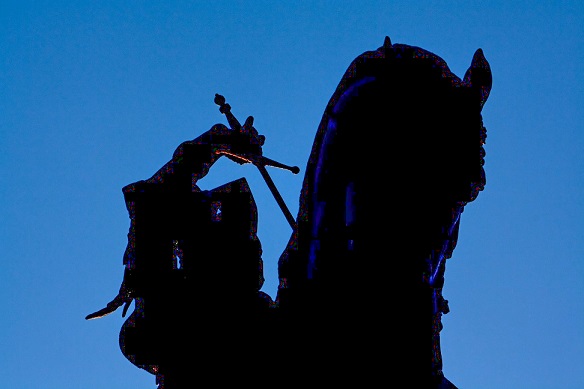}
		\caption{Ours}
	\end{subfigure}
	\caption{Failed case. Our method, as well as other state-of-the-arts, all fail to handle mostly black regions.}
	\vspace{-2mm}
	\label{fig:limitation}
\end{figure}

\begin{figure*}
	\centering
	\includegraphics[height=1.13in]{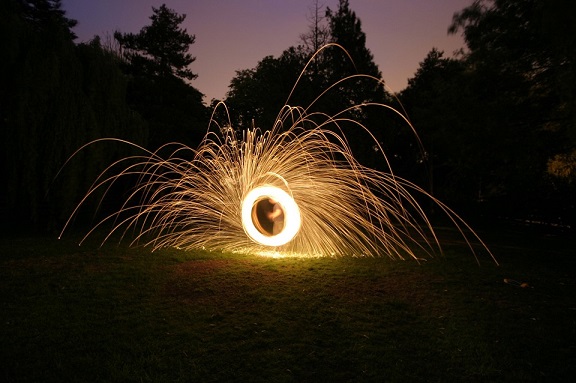}
	\includegraphics[height=1.13in]{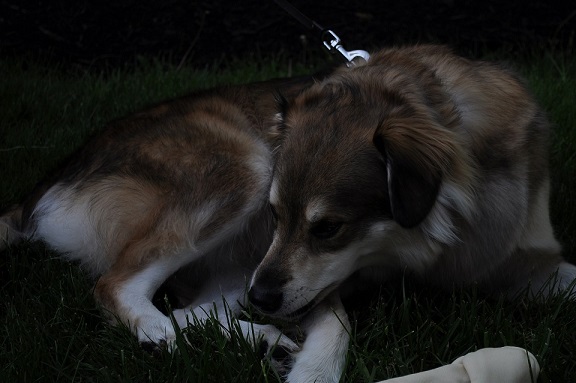}
	\includegraphics[height=1.13in]{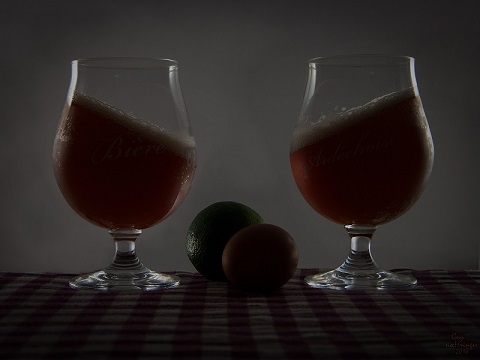}
	\includegraphics[height=1.13in]{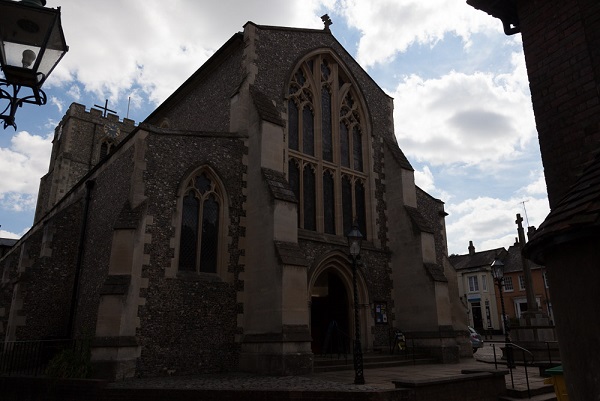}
	\\ \vspace{1mm}
	\includegraphics[height=1.13in]{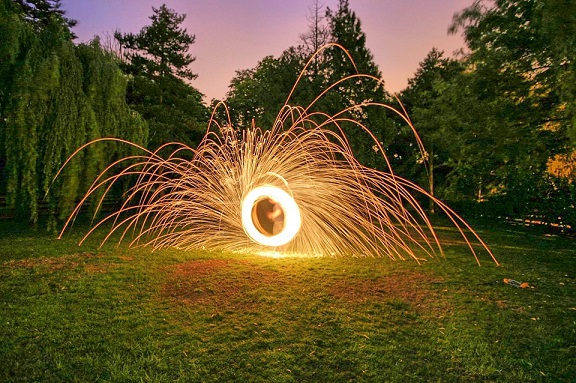}
	\includegraphics[height=1.13in]{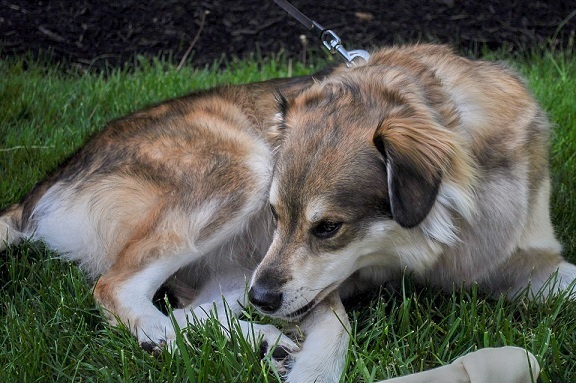}
	\includegraphics[height=1.13in]{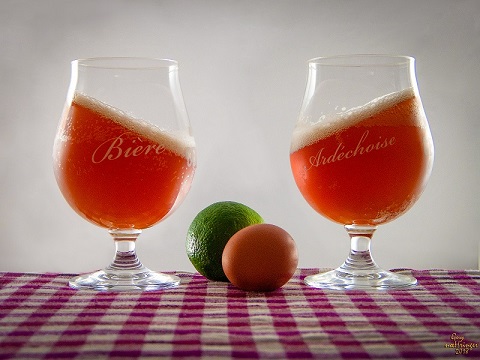}
	\includegraphics[height=1.13in]{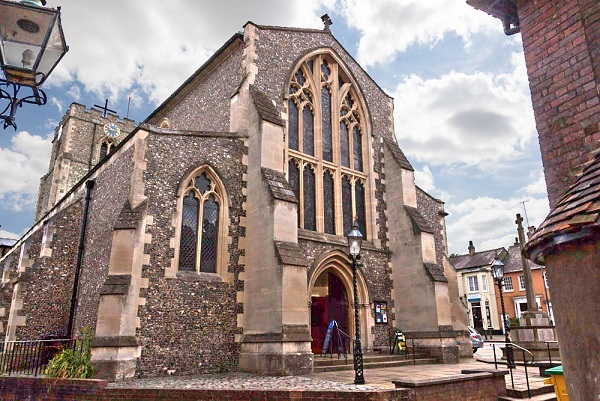}
	\caption{More enhancement results produced by our method. \textbf{Top:} source underexposed images. \textbf{Bottom:} our results.}
	\label{fig:add_res}
	\vspace{-2mm}
\end{figure*}

\vspace{0.3em}
\noindent \textbf{Application to overexposure correction.} Our method is also applicable to overexposure correction. As found by \cite{zhang2019dual}, the inverted version of an overexposed image can be seen as an underexposed image, allowing us to fix overexposed regions by enhancing the corresponding underexposed regions in the inverted image. For a given overexposed image $I$, we first compute its inverted image $\hat{I}$ by $\hat{I} = 1 - I$. Then we perform illumination estimation on $\hat{I}$ to obtain the illumination $\hat{S}$, from which we recover the enhanced image $\hat{R}$. Finally, we get the overexposure corrected result $R$ by performing another inversion operation $R = 1 - \hat{R}$. Fig.~\ref{fig:overexposure} shows two examples.

\vspace{0.3em}
\noindent \textbf{Limitations.} Our method has limitations. As shown in Fig.~\ref{fig:limitation}, our method and the compared state-of-the art methods all fail to produce visually compelling results for the test image in Fig.~\ref{fig:limitation}(a), since the regions of the knight and the horse are almost black and barely have any textures and details. Another limitation is that our method may amplify noise together with the fine scale details when the input image is noisy.

\subsection{Additional Results}
Fig.~\ref{fig:add_res} shows more results produced by our method, where the underexposed images are diverse and involve various lighting conditions, including: (i) a nighttime outdoor image with an irregular light source in the center of the image (1st column), (ii) an evenly exposed image with little details of the dog and the grassland (2nd column), (iii) an indoor image with objects on the desk underexposed (3rd column) and (iv) an unevenly exposed image with the sky normally exposed while the building underexposed (4th column). As shown, for all these cases, our method produces good results.

\section{Conclusion and Future Work} \label{sec:conclusion}
We have presented an approach for enhancing underexposed photos. Unlike previous methods, we reveal the reason why they tend to produce visually unpleasing results from a perspective of perceptual consistency of visual information, and accordingly propose perceptual bidirectional similarity (PBS) for explicitly describing how to maintain the perceptual consistency. Then, we design PBS-constrained illumination estimation for enhancing underexposed photos while avoiding the common visual artifacts. In addition, we extend our method to handle underexposed videos by introducing a probabilistic approach for propagating illumination along the temporal dimension. We have performed extensive experiments on six benchmark datasets, and compared our method with various state-of-the-art methods to demonstrate its superiority.

\section*{Acknowledgment}
The authors would like to thank the anonymous reviewers for their constructive comments. This work was partially supported by the National Key Research and Development Program of China (2016YFB1001001), NSFC (61802453, U1911401, U1811461, 61902275), Fundamental Research Funds for the Central Universities (19lgpy216, D2190670), Guangdong Province Science and Technology Innovation Leading Talents (2016TX03X157), Guangdong NSF Project (2018B030312002, 2019A1515010860), Guangzhou Research Project (201902010037), and Research Projects of Zhejiang Lab (2019KD0AB03). The corresponding author of this work is Wei-Shi Zheng.

\ifCLASSOPTIONcaptionsoff
  \newpage
\fi



%

\bibliographystyle{IEEEtran}
\bibliography{paper.bib}

\begin{IEEEbiography}[{\includegraphics[width=1in,height=1.25in,clip,keepaspectratio]{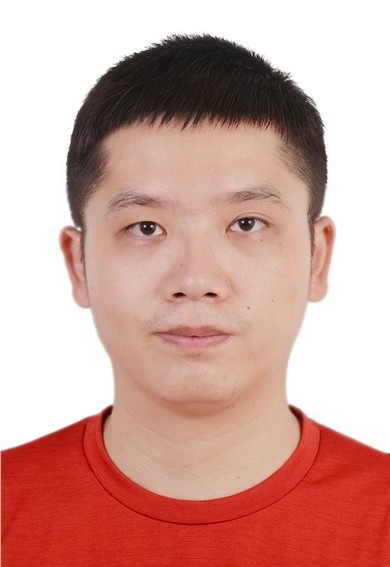}}]{Qing Zhang}
	received his PhD degree in the School of Computer Science from Wuhan
	University in 2017. Currently, he is working as a research associate professor in the School of Data and Computer Science at Sun Yat-Sen University. His research interests include computer graphics, computer vision, and computational photography.
\end{IEEEbiography}

\begin{IEEEbiography}[{\includegraphics[width=1in,height=1.25in,clip,keepaspectratio]{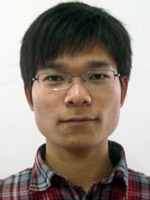}}]{Yongwei Nie}
	received the BSc and PhD degrees from the Computer School of Wuhan University in 2009 and 2015, respectively. Currently, he is an associate researcher at the School of Computer Science \& Engineering, South China University of Technology. His research interests include image and video editing, and computational photography.
\end{IEEEbiography}

\begin{IEEEbiography}[{\includegraphics[width=1in,height=1.25in,clip,keepaspectratio]{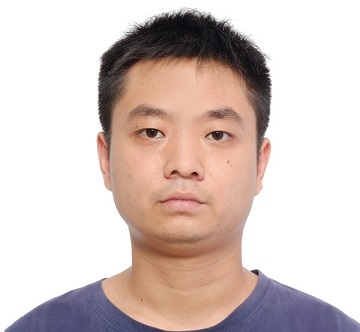}}]{Lei Zhu}
	received his Ph.D. degree in the Department of Computer Science and Engineering from the Chinese University of Hong Kong in 2017.
	He is working as a postdoctoral fellow in the Chinese University of Hong Kong. His research interests include computer graphics, computer vision, medical image processing, and deep learning.
\end{IEEEbiography}

\begin{IEEEbiography}[{\includegraphics[width=1in,height=1.25in,clip,keepaspectratio]{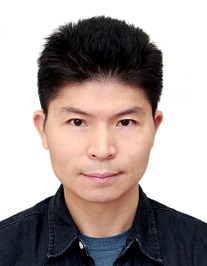}}]{Chunxia Xiao}
	received his BSc and MSc degrees from the Mathematics Department of Hunan Normal University in 1999 and 2002, respectively, and his PhD degree from the State Key Lab of CAD \& CG of Zhejiang University in 2006. Currently, he is a professor in the School of Computer, Wuhan University, China. From October 2006 to April 2007, he worked as a postdoc at the Department of Computer Science and Engineering, Hong Kong University of Science and Technology, and during February 2012 to February 2013, he visited University of California-Davis for one year. His main interests include computer graphics, computer vision and machine learning. He is a member of IEEE.
\end{IEEEbiography}

\begin{IEEEbiography}[{\includegraphics[width=1in,height=1.25in,clip,keepaspectratio]{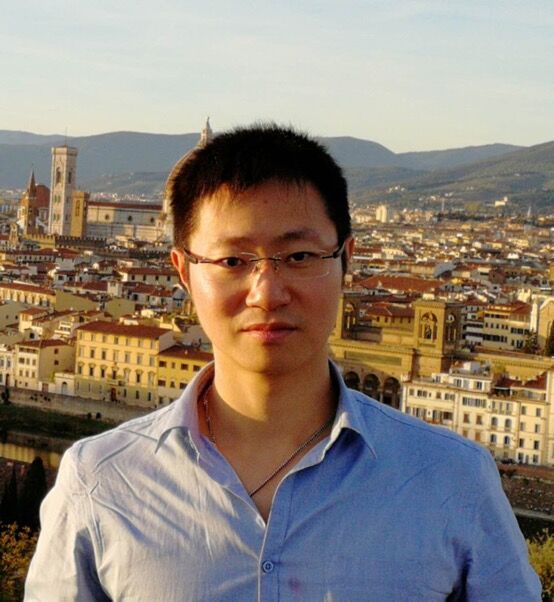}}]{Wei-Shi Zheng}
	received his PhD degree in Applied Mathematics from Sun Yat-sen University in 2008. He is now a full Professor at Sun Yat-sen University. His research interests include person/object association and activity understanding in visual surveillance, and the related large-scale machine learning algorithm. He has now published more than 120 papers, including more than 90 publications in main journals (TPAMI, IJCV, TNN/TNNLS, TIP, PR) and top conferences (ICCV, CVPR, IJCAI, AAAI). He is an associate editor of the Pattern Recognition Journal and area chairs of a number of top conferences. He
has joined Microsoft Research Asia Young Faculty Visiting Programme. He is a recipient of Excellent Young Scientists Fund of the National Natural Science Foundation of China, and a recipient of Royal Society-Newton Advanced Fellowship of United Kingdom.
\end{IEEEbiography}

%

%
%
%




\end{document}